\documentclass[runningheads]{llncs}

\usepackage{eccv}

\usepackage{eccvabbrv}

\usepackage{graphicx}
\usepackage{booktabs}
\usepackage{subcaption}

\usepackage[accsupp]{axessibility}  %

\usepackage[pagebackref,breaklinks,colorlinks,citecolor=eccvblue]{hyperref}

\usepackage{orcidlink}

\usepackage{lipsum}
\usepackage{multicol}
\usepackage{tikz,tkz-kiviat,pgfplots}
\usepackage{xspace}
\usepackage{bm}
\usepackage{multirow} 
\usepackage{makecell}
\usepackage{pifont}
\newcommand{\ours}{Anny\xspace}

\graphicspath{{..}}

\renewcommand{\paragraph}[1]{\vspace{1mm} \noindent \textbf{#1}}
\newcommand{\OurDataset}{Anny-One\xspace}
\definecolor{softgreen}{RGB}{0,150,0} %
\newcommand{\cmark}{\textcolor{softgreen}{\ding{51}}}
\newcommand{\xmark}{\textcolor{red}{\ding{55}}}

\begin{document}

\title{Human Mesh Modeling for Anny Body}
\titlerunning{Human Mesh Modeling for Anny Body}

\author{
Romain Brégier\orcidlink{0000-0002-0554-5857}
\and
Guénolé Fiche\orcidlink{0009-0003-8267-8420}
\and
Laura Bravo-S\'anchez\orcidlink{0000-0003-3556-3391}
\and
Thomas Lucas\orcidlink{0000-0002-6658-6708}
\and
Matthieu Armando\orcidlink{0000-0002-5118-4757}
\and
Philippe Weinzaepfel\orcidlink{0000-0002-4223-3983}
\and
Grégory Rogez\orcidlink{0000-0002-2275-2129}
\and
Fabien Baradel\orcidlink{0000-0003-4625-1713}
}

\authorrunning{R.~Brégier et al.}

\institute{NAVER LABS Europe\\
\url{https://github.com/naver/anny}
}

\maketitle

\noindent\includegraphics[width=1.0\textwidth, trim={0 25px 0 0}]{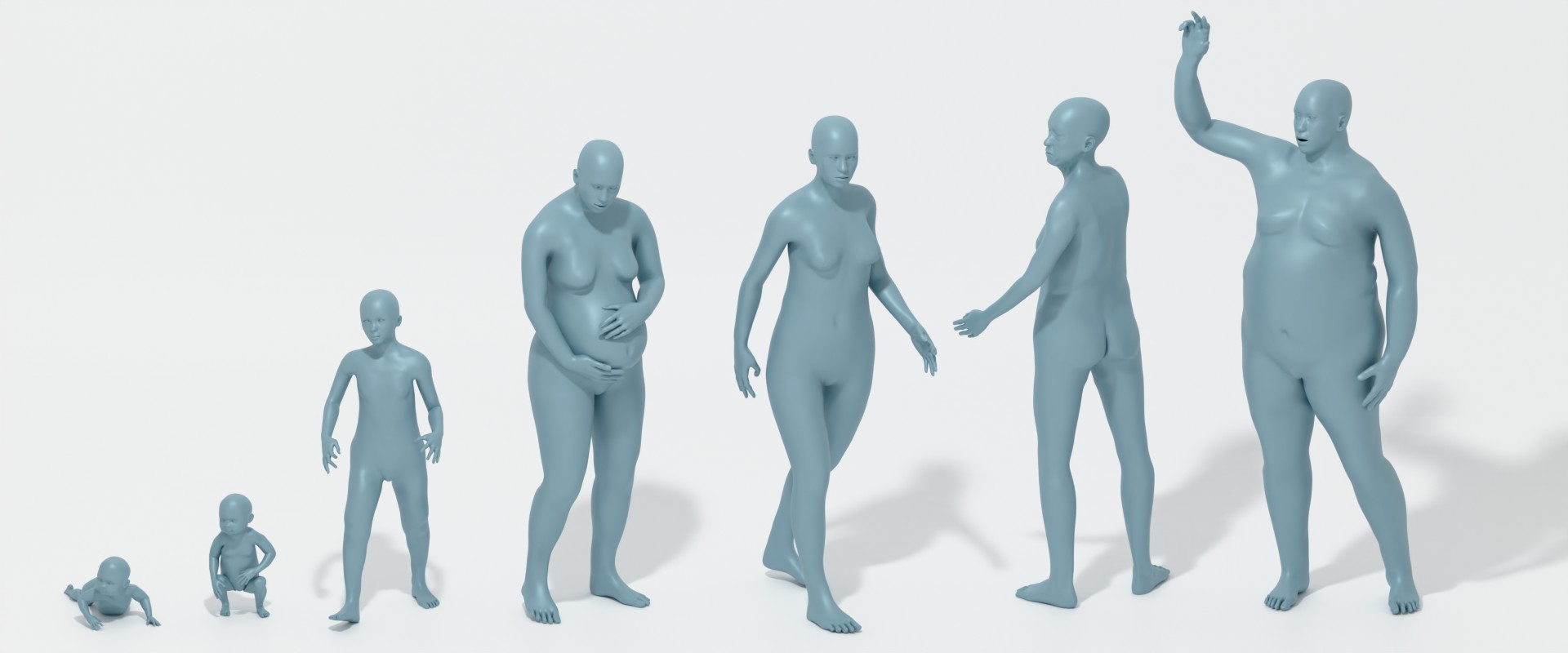}
\captionof{figure}{\textbf{\ours} is a unified, open and interpretable human parametric body model aiming to capture the diversity of human shapes and ages, from infants to elders.\label{fig:teaser}}

\begin{abstract}

Parametric body models provide the structural basis for many human-centric tasks, yet existing models often rely on costly 3D scans and learned shape spaces that are proprietary and demographically narrow.
We introduce \emph{\ours}, a simple, fully differentiable, and scan-free human body model grounded in anthropometric knowledge from the MakeHuman community.
\ours defines a continuous, interpretable shape space, where phenotype parameters (\eg gender, age, height, weight)
control blendshapes spanning a wide range of human forms---across ages (from infants to elders), body types, and proportions.
Calibrated using WHO population statistics, \ours provides realistic and demographically grounded human shape variation within a single unified model.
We release the \ours body model and its code under the \emph{Apache 2.0 license}.
Thanks to its openness and semantic control, \ours serves as a versatile foundation for 3D human modeling---supporting millimeter-accurate scan fitting, controlled synthetic data generation, and Human Mesh Recovery (HMR).
We further introduce \emph{\OurDataset}, a collection of 780k photorealistic images generated with \ours, showing that despite its simplicity, HMR models trained with \ours can match the performance of those trained with scan-based body models.

\end{abstract}

\section{Introduction}

Parametric body models provide a compact and differentiable representation of the human body, enabling a wide range of applications---from scan fitting and animation to Human Mesh Recovery (HMR) from images and videos~\cite{goel2023humans,yin2025smplest,zhang2021pymaf,shin2024wham,kocabas2020vibe}.
Human body models have been central to advances in human-centric perception, enabling tasks such as motion analysis, behavior understanding, and human–robot interaction.
Among them, the SMPL family~\cite{loper2015smpl,pavlakos2019smplx,osman2020star,osman2022supr} has played a transformative role.
Its models achieved compact and accurate modeling capabilities on standard benchmarks by learning low-dimensional shape spaces from 3D body scans~\cite{robinette2002caesar,Bogo:CVPR:2014,Bulat2017FAW}.
However, such data-driven design also comes with limitations:
the available scan datasets are expensive, demographically narrow, and privacy-sensitive.
As a result, %
these models are not suited to
represent the full diversity of human bodies---especially children, elderly individuals, and morphologies uncommon in populations.
Separate models have been proposed for specific demographics, such as SMIL~\cite{hesse2018smil} for infants or distinct male/female SMPL variants, and more recent efforts like SMPL+A~\cite{agora} interpolate between children and adult models.
Yet, obtaining representative data for all body types remains impractical and raises privacy concerns.

Instead of relying on scan data, we explore an alternative foundation for human modeling.
We leverage the anthropometric knowledge embedded in MakeHuman~\cite{makehuman}---a free, community-driven framework designed to model human variability through explicit, interpretable parameters such as age, gender, and body proportions.
This procedural knowledge, built by artists over decades, provides a rich, open description of human morphology that naturally spans diverse body types.
Building on this foundation, we introduce \emph{\ours}, a simple and fully differentiable human body model that replaces learned shape bases with interpretable phenotype parameters.
Each parameter (\eg, age, height, weight, muscle, proportions) is defined in the continuous range [0,1], and directly controls corresponding blendshapes (Figure~\ref{fig:age_interpolation}).
Calibrated using WHO population statistics, \ours covers realistic human variation across the full lifespan---from infants to elders---within a single unified model (Figure~\ref{fig:teaser}).
Because it is built on open assets rather than biometric scans, \ours is free from privacy constraints and can be shared, analyzed, and extended by the community.
Its open-source implementation can be easily integrated into existing pipelines, and allows a wide range of applications within a single unified framework---from millimeter-accurate scan fitting to controllable synthetic data generation and HMR.
To demonstrate its potential, %
we created \OurDataset, a large-scale synthetic dataset of 780k photorealistic images of humans generated with \ours, featuring expressive full-body poses, hands, and faces across diverse environments.
HMR models trained with \OurDataset achieve competitive accuracy on standard benchmarks and outperform existing approaches when body-shape diversity is high.%

Our main contributions are:
(A)	\emph{\ours}, a differentiable, scan-free human body model representing continuous and interpretable shape variations across genders, ages, and body types.
(B)	\emph{\OurDataset}, a large-scale synthetic dataset of 780k photorealistic humans with diverse 3D poses and shapes;
(C)	Empirical validation showing that \ours enables accurate scan fitting and competitive HMR, offering a unified and complete approach to human modeling.

\section{Related work}
\label{sec:related_work}

Modeling humans with 3D parametric meshes has become a central tool in computer vision and graphics, enabling applications ranging from animation to human mesh recovery (HMR) from images and videos. Differentiable body models provide a structured representation of human shape and pose that can be directly integrated with modern vision architectures, allowing efficient inference of full-body geometry from visual observations.%

\paragraph{Parametric body mesh models} represent the human body as a deformable surface mesh controlled by pose and shape parameters.
Important research efforts have been made to build accurate models representative of body surface deformations across different poses and across different individuals characteristics, through data-driven approaches based on 3D human scans.
The SMPL~\cite{loper2015smpl} body model family is arguably the most widely used in the computer vision community. It relies on linear blend shapes to model shape-dependent vertex displacements in rest pose, as well as corrective blend shapes to further refine the output mesh depending on the pose. 
The original model was extended to SMPL-X~\cite{pavlakos2019smplx} to capture full-body meshes, i.e., including facial expressions and hand poses. Other refinements were also considered, in particular STAR~\cite{osman2020star} proposed
a more compact formulation compared to SMPL, and SUPR~\cite{osman2022supr} split the model into individual body parts with a sparse factorization of pose-corrective blend shapes.
To model infants, a separate model was also proposed (SMIL~\cite{hesse2018smil}), and is sometimes combined with SMPL-X through interpolation~\cite{agora}, which we refer to as SMPL-X+A.
Other body models with similar properties have also been introduced, such as GHUM and GHUML~\cite{xu2020ghum} that use the latent space of an auto-encoder as parametrization space.
Concurrent to our work, ATLAS~\cite{park2025atlas} is a model derived from adults scans, that decouples surface and skeletal representations for finer control over body proportions. Ferguson \etal\cite{ferguson2025mhr} latter released a similar model, MHR, with a more restricted rig and an artist-designed facial expression space for easier control.

\paragraph{Modeling shape diversity.}
A major challenge for parametric body models is to capture the diversity of human morphologies.%
The above approaches tackle this by learning shape representations from datasets of 3D scans.
This is hard to achieve because collecting data such as 3D body scans is costly and time consuming.
In particular, the CAESAR~\cite{robinette2002caesar} dataset used to train SMPL comprises people from the USA, Netherlands and Italy aged between 18 and 65 years in early 2000s. It contains fewer than $5000$ individuals, and already constitutes a significant data collection effort.
This limits the diversity that scan-based body models can represent.

\paragraph{Artist-designed human models.}
Instead of relying on large collections of scans, we explore an alternative approach consisting in leveraging artist-designed models.
Accurately modeling the variability of human shape has been the focus of much efforts from computer graphics designers and artists; we leverage these efforts instead of collecting real-world data. 
Various tools exist to design human-like characters~\cite{makehuman, humgen3d, metahuman}. 
We build our work on MakeHuman~\cite{makehuman}, an open-source and community-driven project.
Anny does not rely on expensive collections of 3D scans, yet we find that it is sufficient to achieve competitive results for tasks such as 3D registration and human mesh recovery.

\paragraph{Human mesh recovery} (HMR) aims to reconstruct a full 3D human mesh from a single image or video, as introduced in the eponymous work~\cite{kanazawa2018end}.
Here, we focus on parametric HMR, which refers to methods that output the parameters of a human body model such as SMPL~\cite{loper2015smpl}.
Existing methods can be broadly separated in two categories: single-person and multi-person methods. 
\emph{Single-person} methods assume that human detections or bounding boxes are provided, obtained either from ground truth data or from an off-the-shelf detection model.
In that setting, the HMR model directly regresses body model parameters~\cite{kanazawa2018end}.
Progress was made at the level of architectures and backbones~\cite{goel2023humans,armando2024cross}, camera models~\cite{wang2023zolly,patel2025camerahmr,wang2024blade}, and extensions to expressive body models~\cite{multi-hmr2024,cai2023smpler,lin2023one,sun2024aios,feng2021collaborative}. 
In \emph{multi-person} HMR~\cite{multi-hmr2024,sun2021monocular,sun2024aios,sun2022putting}, all visible humans in the input have to be detected and localized in 3D by the model, in addition to regressing each human mesh.
In that case, the input domain of the model typically shifts from center crops of people to images containing multiple people at arbitrary locations.
Our proposed Anny-One dataset is suitable for both settings, and we empirically evaluate it with both single-person and multi-person state-of-the-art methods.

\paragraph{Training data for HMR.}
Another avenue of research to improve the performance of HMR models is to focus on obtaining better training data. One of the main limiting factors in training HMR models is the lack of images labeled with 3D ground-truth humans. 
While earlier works~\cite{pavlakos2019smplx,bogo2016keep,huang2017towards,lassner2017unite,guler2018densepose} used optimization procedures to fit parametric models to 2D observations such as 2D keypoints, most regression-based models cited above use datasets with 2D~\cite{lin2014microsoft,andriluka14cvpr,johnson2010clustered} or 3D ground-truth joints~\cite{ionescu2013human3,mono-3dhp2017}, sometimes with pseudo-ground-truth meshes~\cite{li2022cliff,joo2021exemplar,kolotouros2019learning,moon2022neuralannot,lin2023one}. Acquiring images with 3D ground truth is expensive, time-consuming, and pseudo-annotations often lack precision in particular for hands and faces. 
In contrast, synthetic datasets offer several advantages: they scale easily, provide noise-free annotations, and allow precise control over appearance, pose, and scene variables.
Existing works have shown that synthetic data can complement~\cite{varol2017learning,agora} or even replace real data to achieve state-of-the-art (SOTA) HMR performance~\cite{bedlam,yin2024whac,multi-hmr2024}, in particular when expressive poses (including hands and faces) are involved~\cite{multi-hmr2024}, or accurate camera estimation in human-centric scenes is  required~\cite{wang2024blade,patel2025camerahmr}.
In this work, we leverage all advantages of synthetic data. 
Our large-scale \OurDataset{} dataset contains diverse humans in terms of body shape and appearance, with expressive poses including faces and hands, with exact ground truth.
We train single-person and multi-person HMR models and show that such large-scale synthetic data, covering a broad distribution of human shapes and poses, can serve as a valuable source of training data, particularly for pretraining HMR models.

\section{Modeling Anny Body}
\label{sec:annybody}

In this paper, we introduce \emph{\ours}, a differentiable parametric mesh model aiming at modeling a large diversity of human morphologies.
\ours is built on assets from \emph{MakeHuman}~\cite{makehuman, mpfb2}, a free and community-driven framework that enables artists to model a wide variety of human-like 3D characters.
We release a PyTorch implementation of \ours under a permissive \emph{Apache 2.0} open-source license to foster the development of human-centric research and applications.

\begin{figure}[!b]
\centering\small
\hspace{-8mm}
\includegraphics[height=3.7cm]{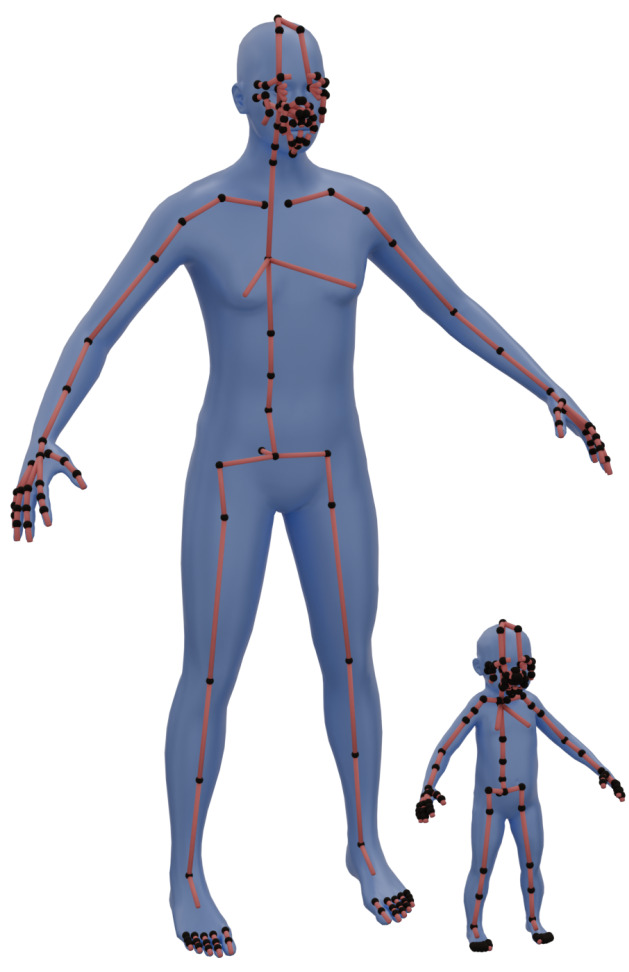}
\vrule
\includegraphics[height=3.7cm]{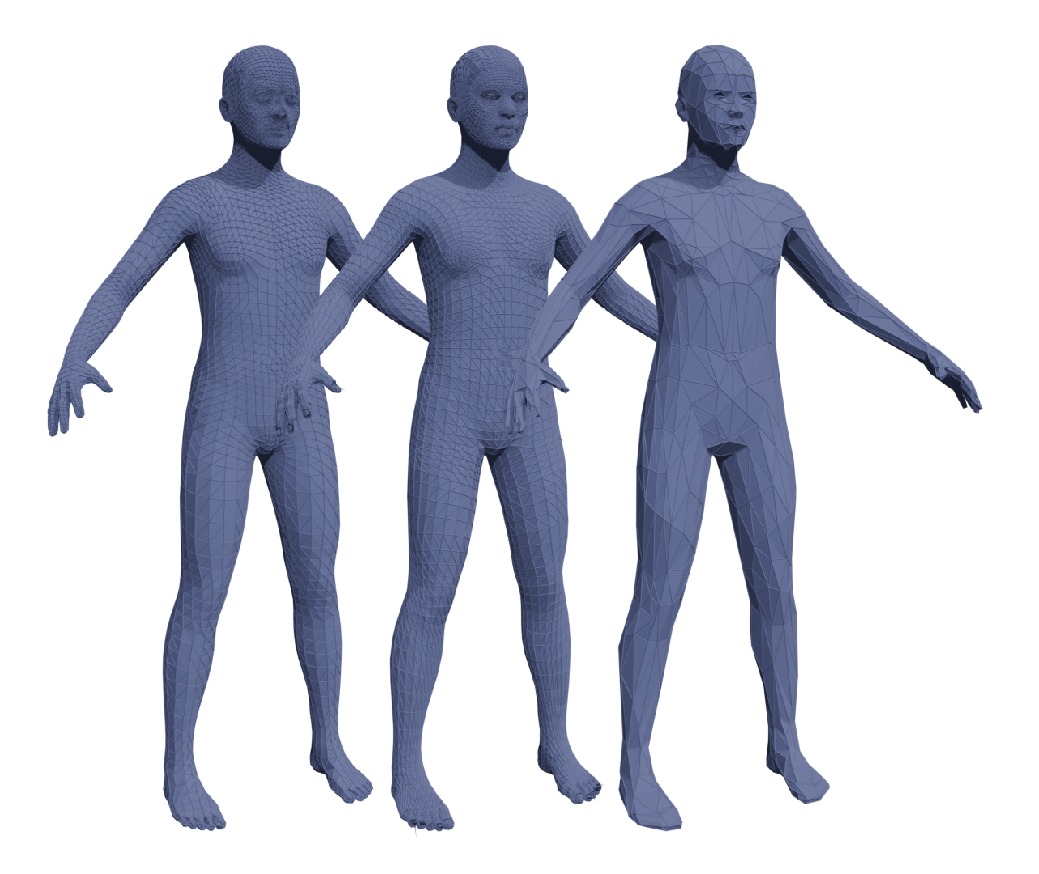}
\vrule
\includegraphics[height=3.7cm]{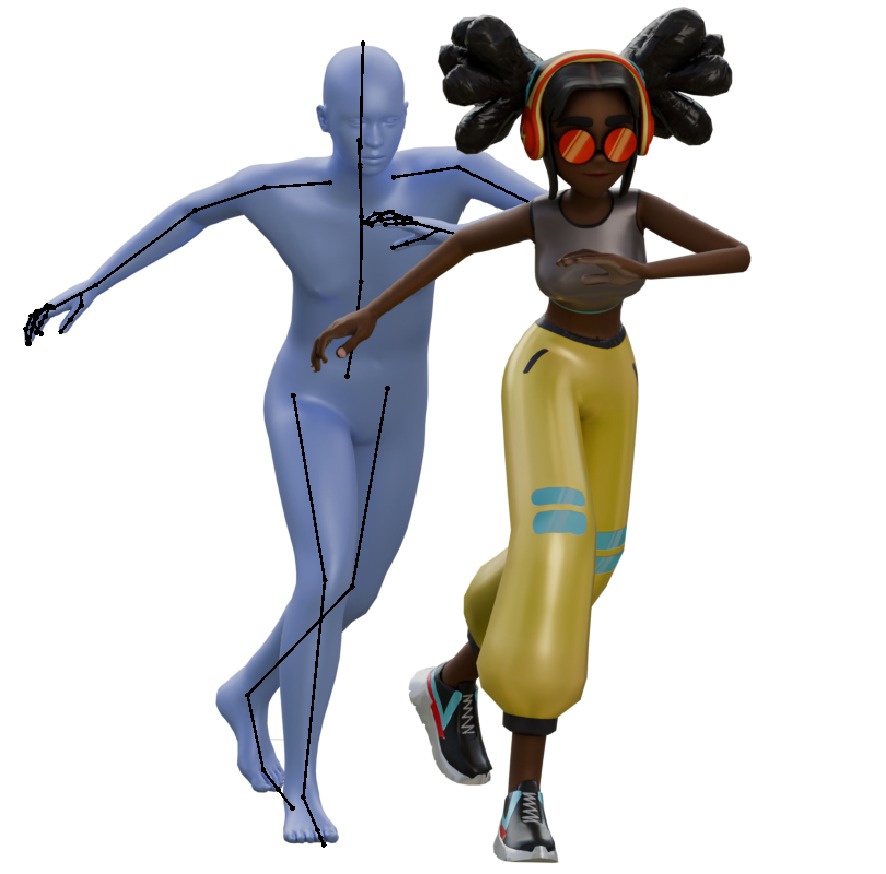}
\hspace{-12mm}
\vspace{-0.2cm}
\caption{\label{fig:skeleton}\textbf{Mesh and skeleton of Anny}. \emph{Left:} default mesh and skeleton, modeling two different morphologies. 
\emph{Middle:} different mesh topologies (\emph{default}: 13,718 vertices---\emph{SMPL-X}: 10,475 vertices---\emph{coarse}: 1,229 vertices).
\emph{Right:} Anny and Mixamo~\cite{mixamo} character using the same skeleton and bone orientations.}
\end{figure}

\paragraph{Base model.}
The default mesh of Anny is composed of $V = 13{,}718$ vertices and 13,710 quadrilateral faces (including tongue and eyes). Vertices are softly attached to a skeletal rig of $B = 163$ bones, illustrated in Fig.~\ref{fig:skeleton}, using $B \times V$ sparse skinning weights.
To ensure compatibility with the MakeHuman ecosystem, Anny closely follows MakeHuman specifications, with minor edits to remove sensitive anatomical details such as nipples and genitals, and to ensure left/right skinning symmetry.
Anny also supports alternative mesh topologies and rigs, which allows simple interoperability and character animation re-targeting.%

\begin{figure}[!t]

    \centering
    \small
    \noindent
    \begin{minipage}[b]{0.55\linewidth}
    \includegraphics[width=\linewidth]{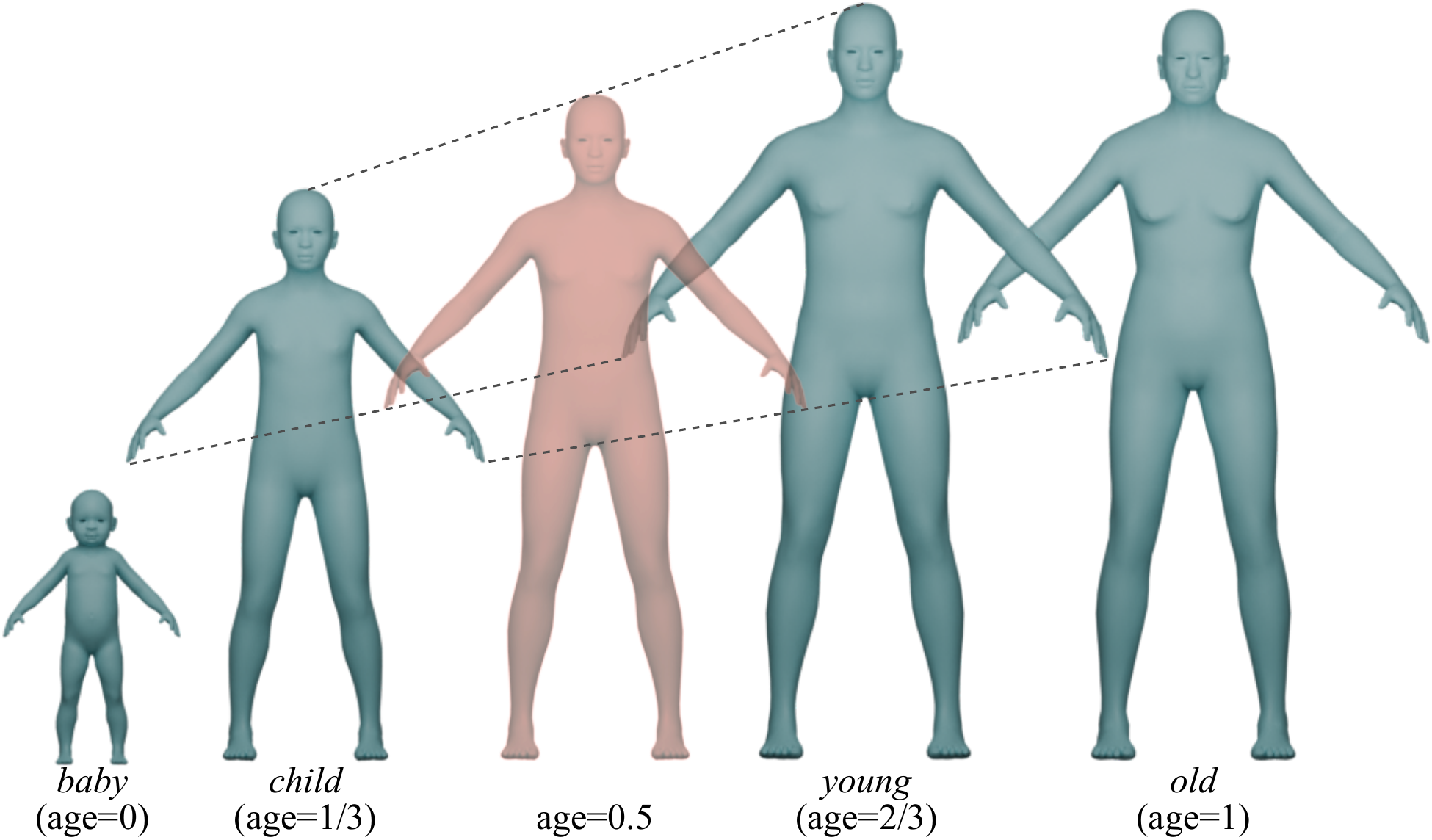}
    \end{minipage}
    \begin{minipage}[b]{0.43\linewidth}
        \centering
        \begin{minipage}[b]{\linewidth}
        \centering
        \includegraphics[height=1.2cm]{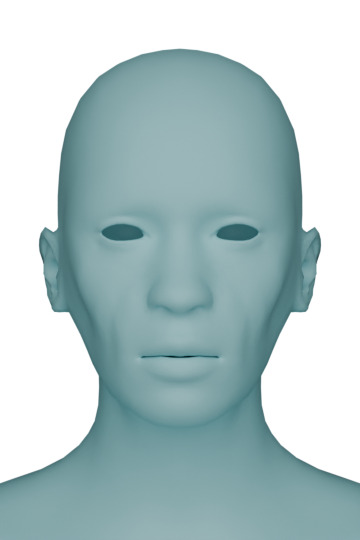}
        \includegraphics[height=1.2cm]{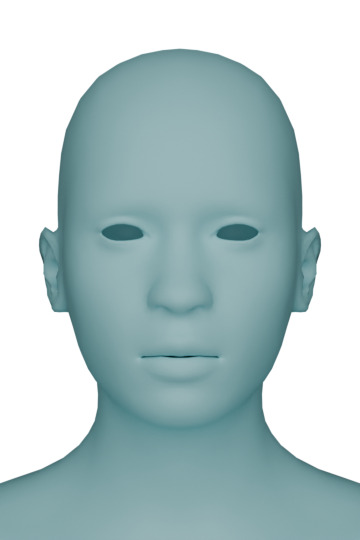}
        \includegraphics[height=1.2cm]{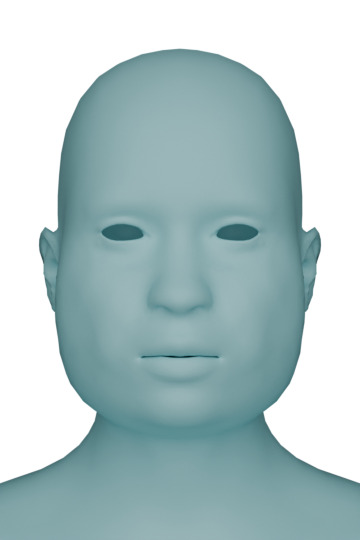}
        \scriptsize
        
        \emph{head-fat}
        \end{minipage}
        
        \begin{minipage}[b]{0.49\linewidth}
        \centering
        \hspace{-0.8cm}
        \includegraphics[height=2.2cm]{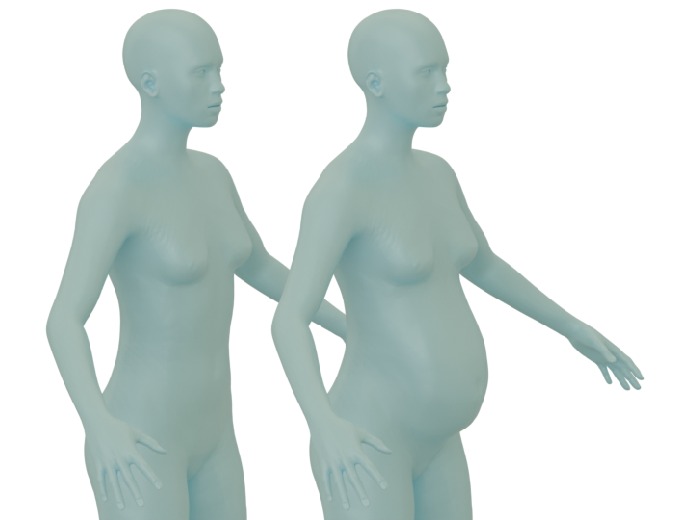}
        \hspace{-0.8cm}
        
        \scriptsize
        \emph{stomach-pregnant}
        \end{minipage}
        \begin{minipage}[b]{0.49\linewidth}
        \centering
        \hspace{-0.8cm}
        \includegraphics[height=2.2cm]{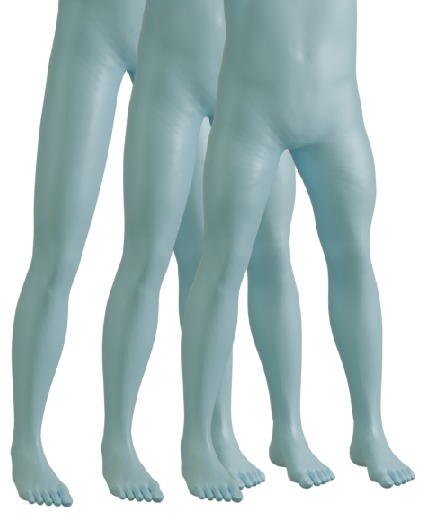}
        \hspace{-0.8cm}
        
        \scriptsize
        \emph{upperleg-height}
        \end{minipage}
    \end{minipage}

    \vspace{-2mm}

    \caption{\textbf{Shape parametrization} is implemented using piece-wise multi-linear interpolation between prototypical shape. Left: illustration with \emph{age}. Right: example of local morphological variations covered by the model.}
    \label{fig:age_interpolation}
    \label{fig:local_changes}
    \vspace{-6mm}
\end{figure}

\paragraph{Phenotypes.}
We capture the diversity of human morphologies through various parameters which we refer to as \emph{phenotypes}, following MakeHuman terminology. These phenotypes aim to encode high-level characteristics, such as \emph{age}, \emph{gender}, \emph{weight}, \emph{muscle} amount, \etc, as well as more local changes, such as the amount of \emph{head fat} or changes in belly morphology during pregnancy  (Fig.~\ref{fig:local_changes}).
Specifically, artists created 1,136 prototypical mesh blendshape variations, each corresponding to a particular combination of phenotype attributes. We extend these discrete variations into a continuous shape space by piece-wise multi-linear interpolation of vertex positions.

\paragraph{Word of caution.}
Phenotypes are based on preconceptions of artists regarding particular human traits. As a result, they encode by design stereotypes of MakeHuman artists, and \textbf{\textit{one should not expect phenotype parameters to faithfully encode any identity-related characteristics, such as gender, age or ethnicity}}.
We nonetheless found these parameters useful to model the diversity of human morphologies, and we keep MakeHuman terminology to make explicit the artists' intent.
Existing scan-based models~\cite{loper2015smpl, pavlakos2019smplx, osman2020star, xu2020ghum, osman2022supr,park2025atlas, ferguson2025mhr} avoid addressing this delicate question by relying on abstract shape spaces with no explicit semantics---except for gender-specific models. It is worth mentioning that these models also convey implicit biases, related to their topology (all models assume individuals with four limbs) and to their training distribution (most models are only designed for adults, for example).

\newcommand{\captionnedheadgraphics}[2]{
    \begin{minipage}[t]{0.15\linewidth}
    \centering
    \scriptsize
    \includegraphics[height=0.70in]{#1}
    
    \emph{#2}
    \end{minipage}}

\newcommand{\captionnedfootgraphics}[2]{
    \begin{minipage}[t]{0.16\linewidth}
    \centering
    \scriptsize
    \includegraphics[height=0.5in]{#1}
    
    \emph{#2}
    \end{minipage}}
    
\paragraph{Differentiable deformation.}
Anny takes shape and pose parameters as input and outputs a 3D posed human mesh representation in a backward-differentiable manner.
The shape of the mesh is controlled by scalar coefficients targeting various phenotypes regarding age, weight, gender, \etc.
These coefficients define weights of a piecewise-multilinear interpolation between prototypical blendshapes, that are used to adjust the shape of a base model.
For instance, blendshapes corresponding to the phenotypes \emph{child} ($age{=}1/3$) and \emph{young} ($age{=}2/3$) contribute equally to the mesh deformation given a parameter value $age{=}0.5$, as illustrated in ~Fig.~\ref{fig:age_interpolation}.
This use of interpolation constraints the structure of the shape space, and helps producing topologically consistent meshes.
The shape-adjusted model features a skeleton, composed of a set of bones connected along a kinematic tree. One bone is defined as root bone of the tree.
To encode a pose for the model, the pose of the root bone is given, together with 3D rotations at each joint between connected bones, relative to their rest configuration.
Given pose parameters, we apply forward kinematics to retrieve bone poses.
We then deform the mesh using blend skinning, producing the final human mesh, shaped and posed accurately according to the input parameters.
These steps are implemented using PyTorch~\cite{paszke2017automatic} and NVIDIA Warp~\cite{warp2022} to benefit from automatic gradient back-propagation features of these libraries.

\paragraph{Self-intersection.} To prevent self-intersection, we test for intersecting faces belonging to different body parts---using a bounding volume hierarchy for computational efficiency---and discard corresponding meshes during data generation.

\paragraph{Interoperability.}
To ensure interoperability with previous work, we define mappings between Anny's default topology and existing mesh body models. These mappings serve mainly two purposes. The first is to empirically evaluate models trained with Anny on existing benchmarks; for that purpose we define a mapping to regress SMPL-X vertices from Anny meshes and vice versa. The second one is to generate synthetic 3D scenes with Anny annotations. Our synthetic data generation pipeline is built on Humgen3D~\cite{humgen3d}, thus we also learn regressors for this body model.
More specifically, we optimize sparse linear regressors $\bm{R} \in \mathbb{R}^{M \times N}$ to map vertex coordinates $\bm{v}_j \in (\mathbb{R}^3)^N$ from a first body model to coordinates $\hat{\bm{v}}_i = \sum_j \bm{R}_{i,j} \bm{v}_j \in (\mathbb{R}^3)^M$ of a different body model with a different topology.
This is achieved by first fitting Anny to a set of meshes with the target connectivity (\eg SMPL-X), and by initializing the regression coefficients $\bm{R}$ as the barycentric coordinates of the projection of target mesh vertices onto the source mesh. Coefficients of $\bm{R}$ are then refined jointly together with Anny parameters to minimize the mesh-to-mesh distance, while enforcing the left/right symmetry of the mapping. Applying the direct and reverse mappings between SMPL-X and Anny leads to a $1.3mm$ mean cyclic consistency error (resp. $1.7mm$ between Anny and Humgen3D).

\section{Shape statistical modeling}

A model capturing the distribution of human morphologies can be useful for applications such as data synthesis or as a prior in optimization.
Existing body models~\cite{loper2015smpl, park2025atlas} typically feature a latent space encoding a shape distribution derived from their training sets, \eg using PCA for SMPL. These training sets consist of proprietary collections of 3D scans which required significant acquisition effort, yet they remain insufficient to represent the global population. Datasets such as CAESAR~\cite{robinette2002caesar} and SizeUSA~\cite{sizeusa} consist of scans of adults from industrialized countries, and as such are not representative of morphological variations among the global population and all ages.
As a consequence, a normal sampling of the SMPL-X neutral shape distribution produce rather tall and overweight bodies (average height: 172$cm$, average body mass index: $25.5kg/m^2$, assuming a constant buoyancy of 0.98) compared to global adult statistics (2019 estimate for 19-years old girls/boys: 158.6$cm$/170.8$cm$~\cite{rodriguez2020height}).

\begin{figure*}
	\centering
    \vspace{-3mm}
    \includegraphics[width=\linewidth]{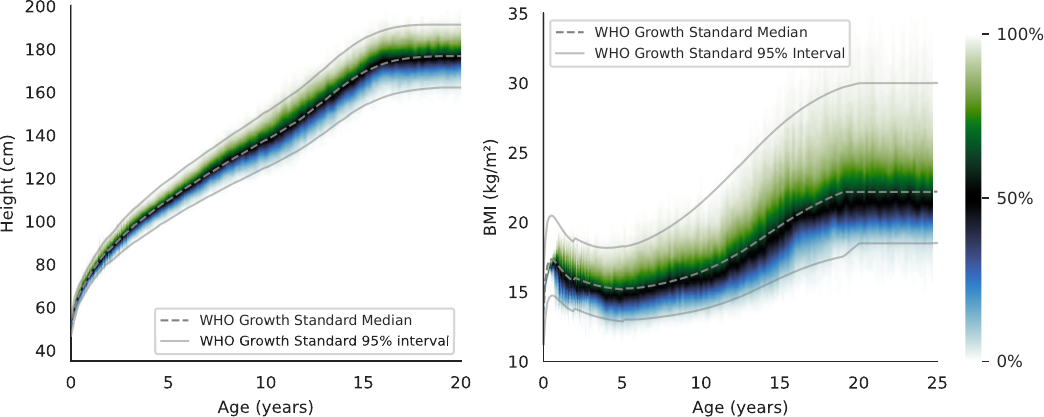}
    \vspace{-6mm}
	\caption{\textbf{Statistical shape modeling.}
    We calibrate the Anny shape distribution to match the WHO Child Growth standards for \emph{height-for-age} and \emph{Body Mass Index-for-age}~\cite{who2006growth} (curves for boys).}
	\label{fig:anny_who_calibration_left}
    \vspace{-6mm}
\end{figure*}

\begin{figure*}
	\centering
    \includegraphics[width=\linewidth]{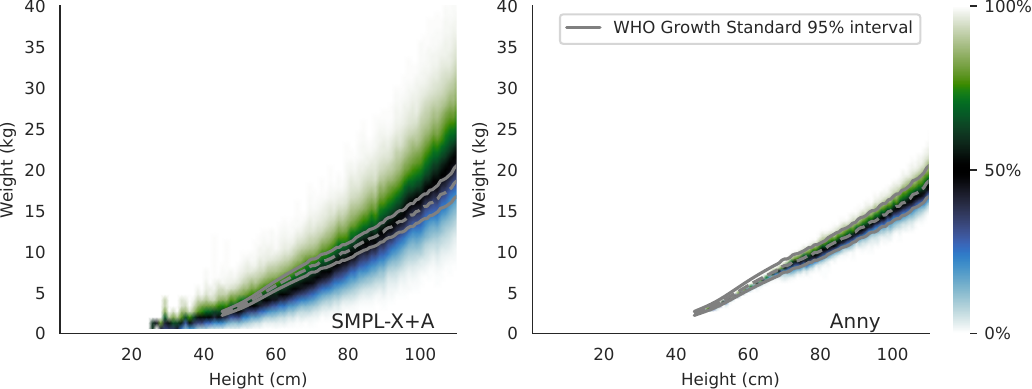}
    \vspace{-6mm}
	\caption{\textbf{Statistical shape modeling (weight–height).}
    The calibrated \ours shape distribution better matches WHO \emph{weight-for-height} growth standards (mixed gender) than the SMPL-X+A shape space.}
	\label{fig:anny_who_calibration_right}
    \vspace{-6mm}
\end{figure*}

Anny can model a large diversity of morphologies through its various phenotype parameters. This diversity covers common body shapes, but also uncommon ones such as 2.4 meters tall individuals.
Human morphology is highly diverse and characterizing the plausibility of Anny body shapes is challenging. Nonetheless, we can
calibrate the distribution of Anny parameters to obtain a shape distribution that is more representative of the global population.
We empirically define a bijective mapping between the \emph{age} parameter of Anny and some morphological age in years. We then model distributions for major phenotype parameters as Beta distributions, conditioned on age and gender. These distributions are jointly calibrated to match the mean and standard deviation of reference growth standards for height and body mass index from the World Health Organization (WHO)~\cite{who2006growth}, shown in Fig.~\ref{fig:anny_who_calibration_left}.
After calibration, the resulting distribution better matches weight-for-height reference data than the SMPL-X+A shape distribution, as shown in Figure~\ref{fig:anny_who_calibration_right}.

\section{\label{sec:fitting}Assessing 3D modeling capabilities}

\begin{figure}[!t]
    \scriptsize
    \noindent
    \begin{minipage}[c]{0.9\linewidth}
    \includegraphics[width=1.111111\linewidth,trim={10px 300px 10px 250px},clip]{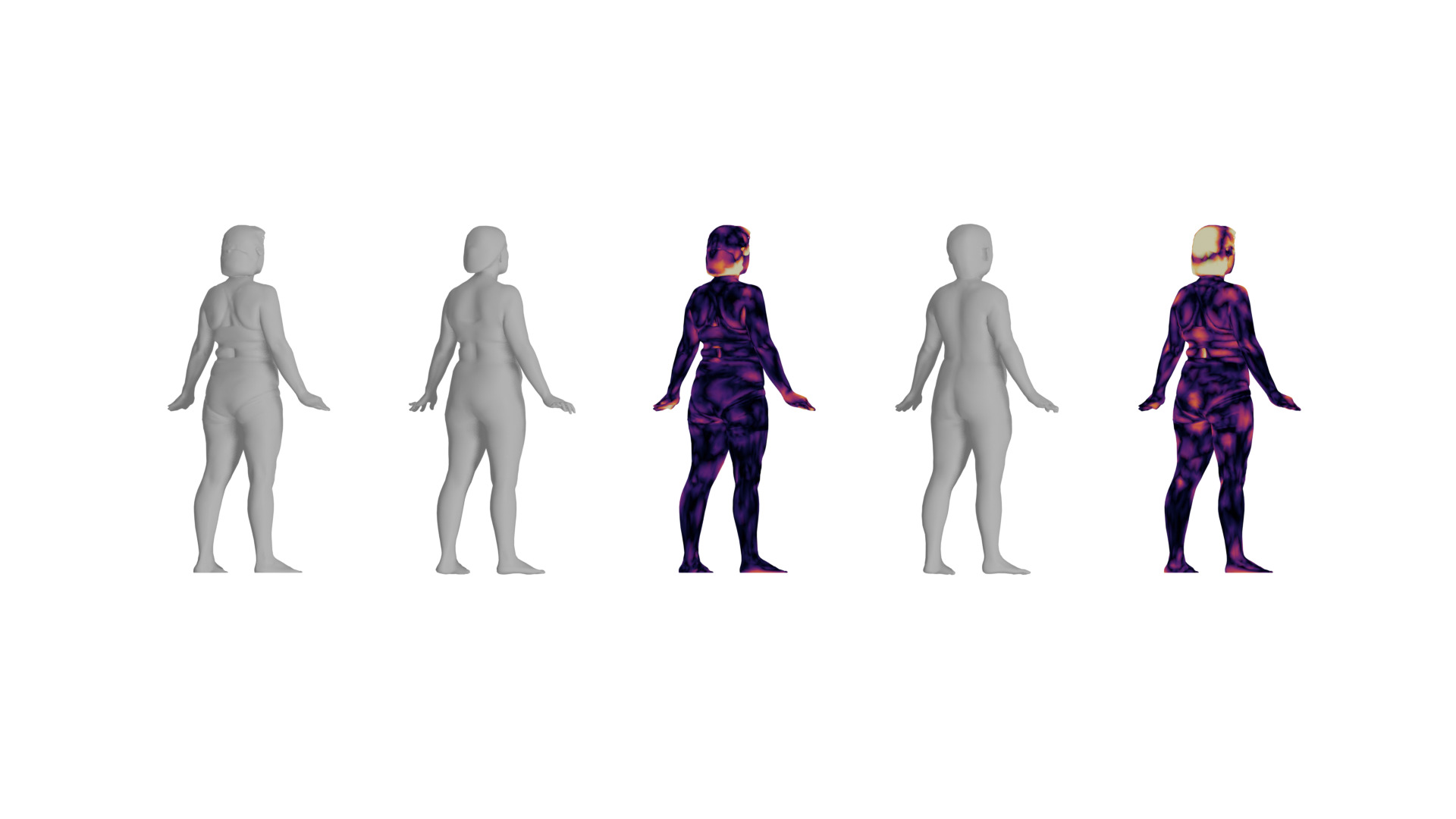}%
    \hspace{-0.1111111\linewidth}%
    \end{minipage}
    \begin{minipage}[c]{0.05\linewidth}
    \centering
    \scriptsize
    15mm
    
    \includegraphics[width=4mm, height=12mm]{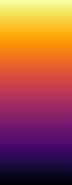}

    0mm    
    \end{minipage}
    
    \noindent\includegraphics[width=\linewidth,trim={10px 300px 10px 250px},clip]{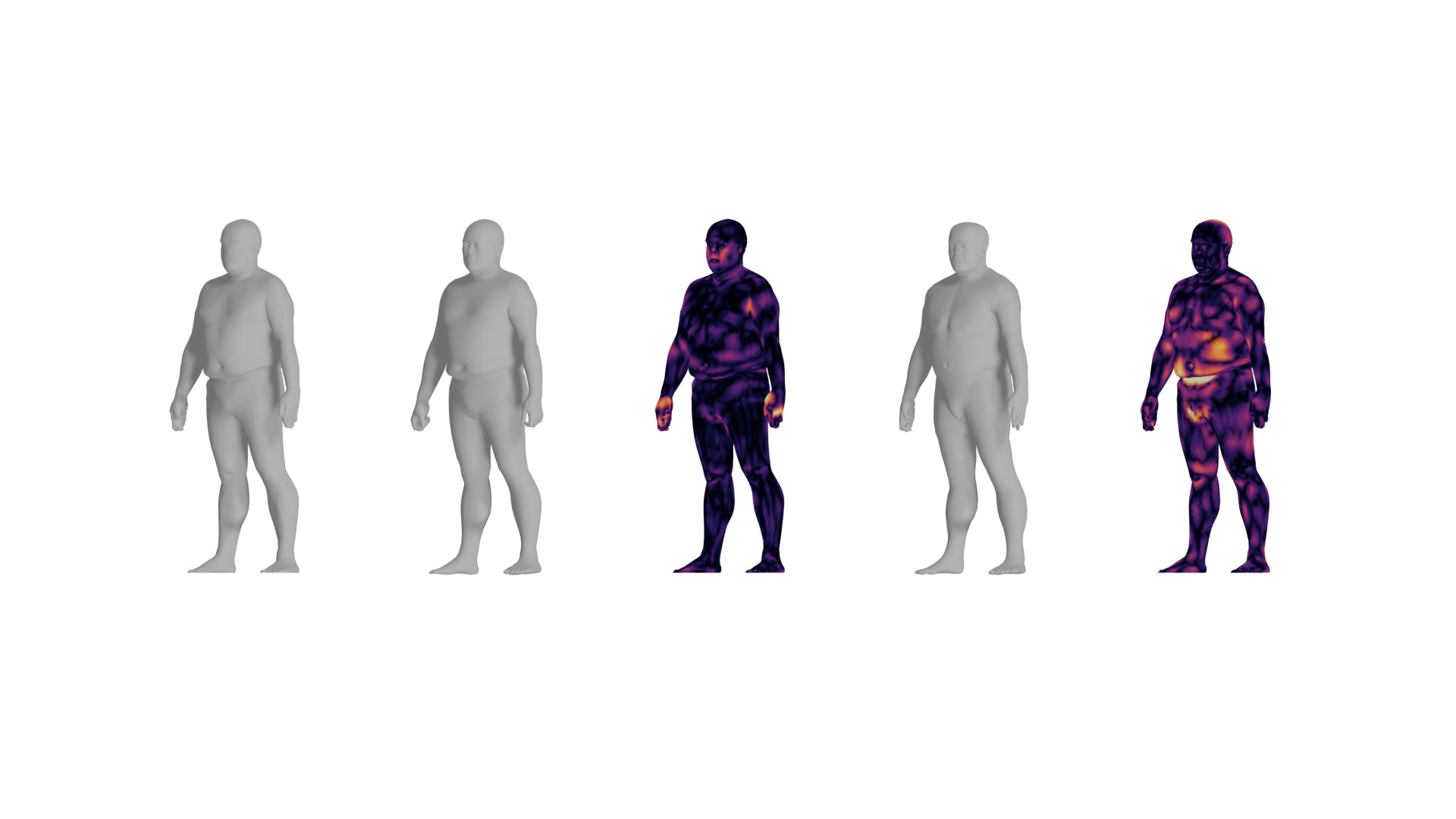}
    
    \noindent\hspace{1.5cm} \textbf{Scan} \hspace{0.8cm} \textbf{SMPL-X (300 components)} \hspace{0.3cm} \textbf{Anny (164 components)}

    \noindent\includegraphics[width=\linewidth,trim={10px 100px 10px 550px},clip]{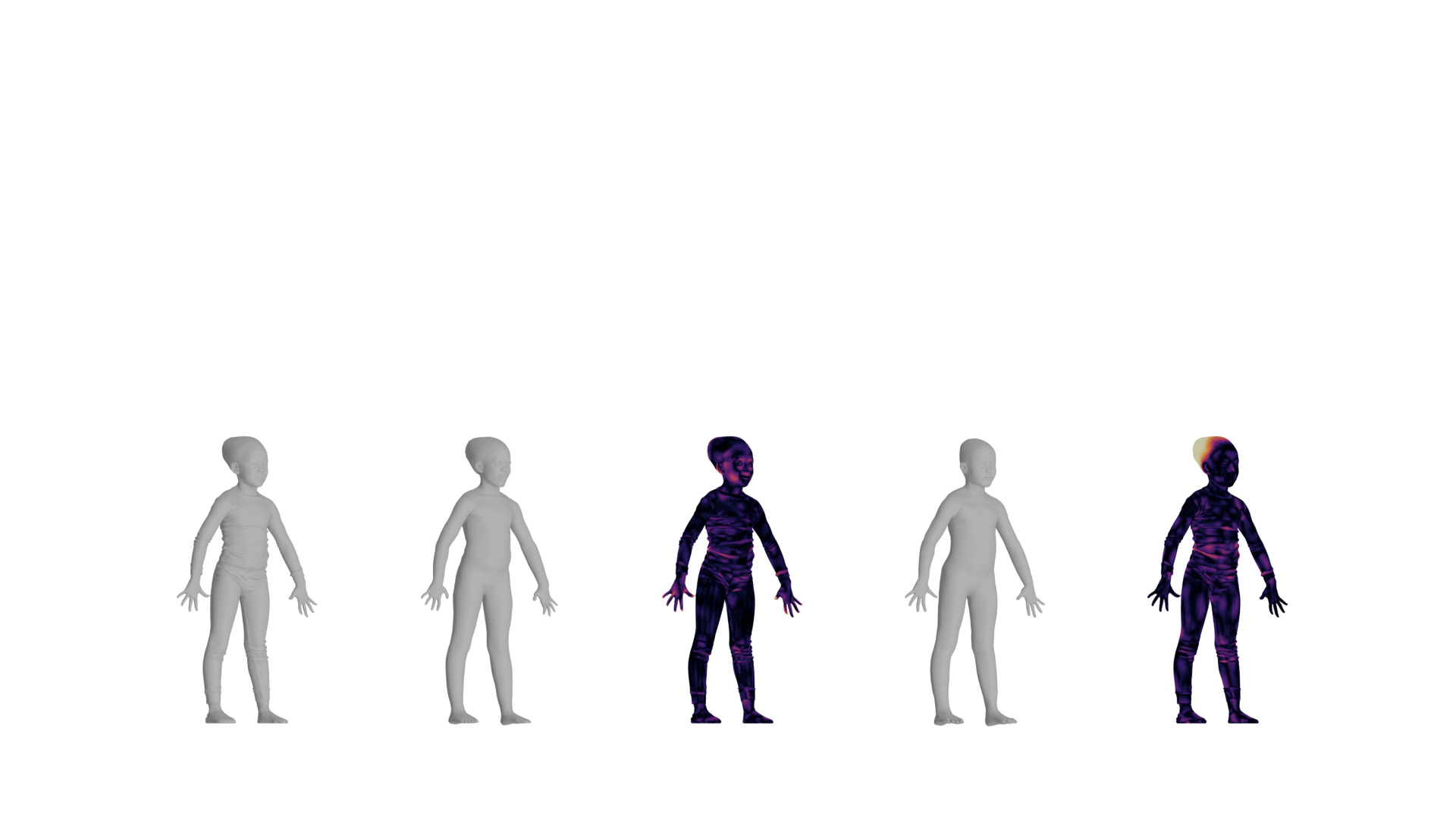}
    
    \noindent\hspace{1.5cm} \textbf{Scan} \hspace{0.8cm} \textbf{SMPL-X+A (301 components)} \hspace{0.3cm} \textbf{Anny (164 components)}
    
    \vspace{-0.2cm}
    \caption{\textbf{3D Registration} of body models to scans of adults and children. Anny can closely approximate 3D scans, but does not model deformations related to clothing or hair contrary to existing scan-based models.\label{fig:3dregistration}}
    \label{fig:placeholder}
    \vspace{-6mm}
\end{figure}

\paragraph{Adults}. To evaluate the expressiveness of the proposed body model, we register Anny to scans from 3DBodyTex~\cite{saint20183dbodytex}, which contains 400 scans of adults (100 male and 100 female individuals) in minimal clothing.
As shown in Fig.~\ref{fig:3dregistration}, Anny produces visually accurate registrations that closely follow the scanned geometry accross subjects.
Quantitatively, it achieves an average scan point-to-mesh error of 
2.5$mm$.
Following established evaluation protocols~\cite{osman2022supr,park2025atlas}, we exclude the head and hands from the quantitative analysis due to lower scan quality in these regions.
Smaller fitting errors have been reported with scan-based models, 
notably a 1.8$mm$ mean error using ATLAS ~\cite{park2025atlas}. We achieved a similar error of 1.9$mm$ using SMPL-X with 300 shape components.
Qualitatively, we observe that SMPL-X better captures fine skin-fold details of high-BMI bodies compared with Anny. However, we also observe that SMPL-X overfits to mesh deformations due to hair and clothings, since scan-based models are specifically designed to capture the geometry of individuals wearing minimal clothes.
In contrast, Anny models neither clothes nor the shape deformation they cause, yet is able to approximate 3D scans with competitive accuracy.

\paragraph{Children}. No public child-scan dataset is available to our knowledge, which motivates Anny's scan-free design.
To assess modeling capabilities with children, we register Anny to three commercial scans from RenderPeople (\emph{posed 320 26}, \emph{48 26} and \emph{502 26}). Although this should not be interpreted as a comprehensive benchmark, Anny achieves lower average scan-point-to-mesh error than SMPL-X on all three scans, with 2.0 \vs 3.2$mm$ mean error.
Together with the AGORA-Kids results in Table~\ref{tab:shape}, these results provide supporting evidence that Anny can competitively model child body shapes.

\begin{figure}
    \centering
    \includegraphics[
      width=0.4\linewidth,
      trim={0 4mm 0 4mm},
      clip
    ]{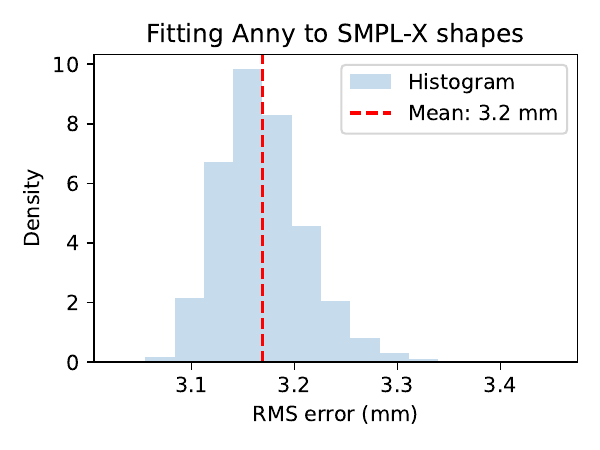}
    \includegraphics[
      width=0.4\linewidth,
      trim={0 4mm 0 4mm},
      clip
    ]{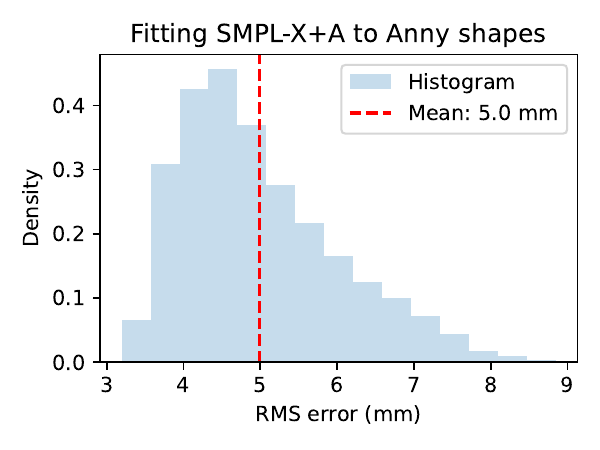}
    \caption{\textbf{Histogram of registration errors} of the Anny model (resp. SMPL-X+A) onto meshes uniformly sampled within the SMPL-X (resp. Anny) shape parameter space. Results suggest that Anny covers a broader shape diversity than SMPL-X+A.
    \label{fig:smplx_anny_registrations}}
\end{figure}

\paragraph{Shape diversity}.
To compare the diversity of shapes modeled by Anny with prior work, we register the SMPL-X+A model to 8000 meshes sampled with uniform Anny shape parameters,
and we register the Anny model to 8000 meshes sampled with uniform SMPL-X shape parameters (within a ball of 2 standard deviation radius~\cite{smpl-made-simple}, and 300 PCA coefficients.).
All meshes are in rest pose using a common SMPL-X topology, thanks to the interoperability mapping presented in Sec.~\ref{sec:annybody}.
We report the histograms of registration errors in Fig.~\ref{fig:smplx_anny_registrations}.
We observe that meshes from the SMPL-X shape distribution admit in general a close Anny approximation (3.2$mm$ mean RMS error), whereas a thick tail of Anny meshes have no close SMPL-X+A equivalent (5.0$mm$ mean RMS error). This suggests that the shape space of Anny covers a larger diversity of morphologies than the one of SMPL-X.
A qualitative analysis of the worst registration results notably %
suggests that it is difficult to model some very tall individuals using the SMPL-X+A shape space.

\section{Recovering \ours Human Meshes}

Human Mesh Recovery (HMR) is a natural downstream task for assessing the representational power of a body model~\cite{multi-hmr2024,patel2025camerahmr}.
Recent works~\cite{bedlam,tesch2025bedlam2} have shown that large-scale synthetic data can be as effective as, or even superior to, real data---whose annotations often contain inherent noise---for training HMR models.
Having a body model interoperable with models that can accommodate clothing furthermore enables to easily generate large-scale, diverse synthetic training data.
In this paper, we leverage these properties and introduce \emph{\OurDataset}, a synthetic dataset of images with corresponding Anny annotations (Section~\ref{sub:hmrdata}).
We then use \OurDataset\ to train HMR models, that we evaluate on standard benchmarks (Section~\ref{sub:hmrmodel}).

\subsection{The \OurDataset dataset}
\label{sub:hmrdata}

Our synthetic dataset, \OurDataset, contains 780k images designed for training and evaluating Human Mesh Recovery models.
It features realistic human meshes with diverse body shapes, poses, and appearances, situated in rich and varied scene contexts.

\begin{figure}[!t]%
    \newlength{\annyonefigwidth}
    \setlength{\annyonefigwidth}{0.18\linewidth}
    \centering
    \includegraphics[width=\annyonefigwidth]{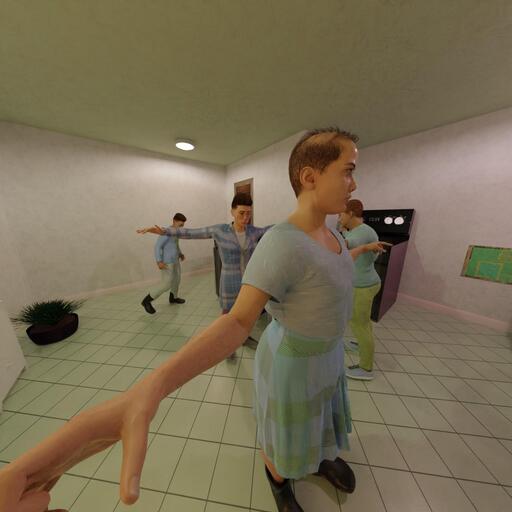}%
    \includegraphics[width=\annyonefigwidth]{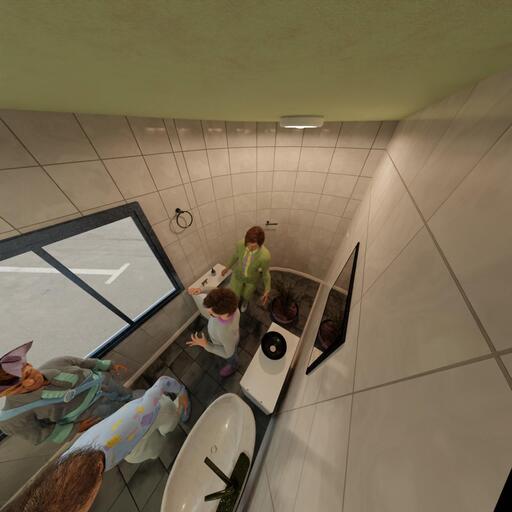}%
    \includegraphics[width=\annyonefigwidth]{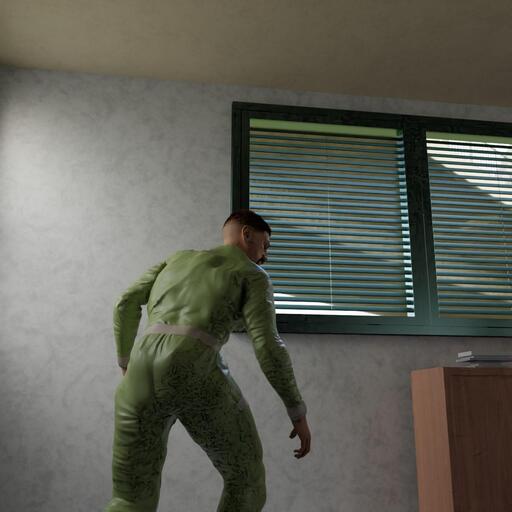}%
    \includegraphics[width=\annyonefigwidth]{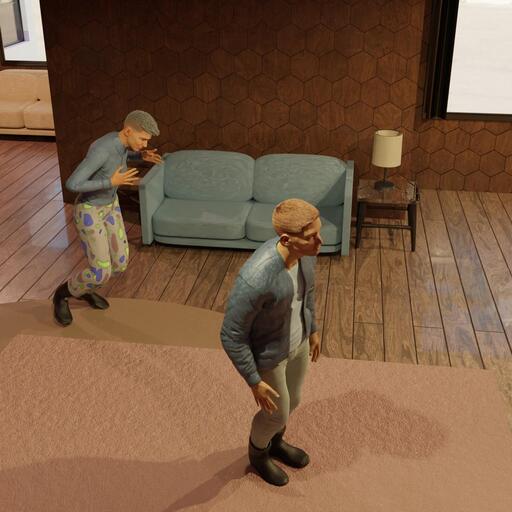}%
    \includegraphics[width=\annyonefigwidth]{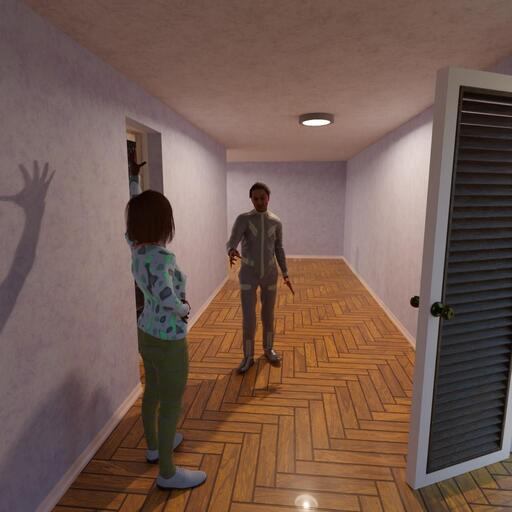}%

    \vspace{-1pt}
    \includegraphics[width=\annyonefigwidth]{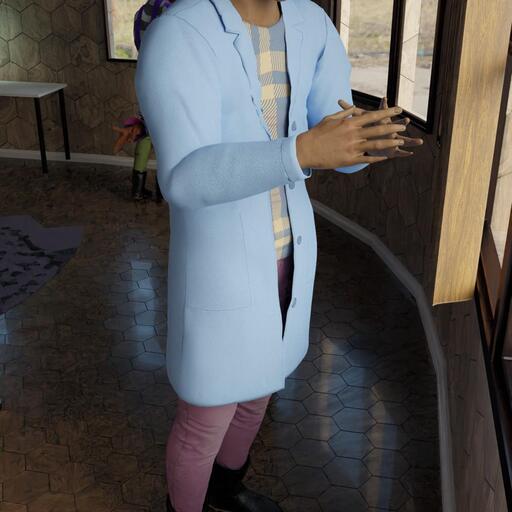}%
    \includegraphics[width=\annyonefigwidth]{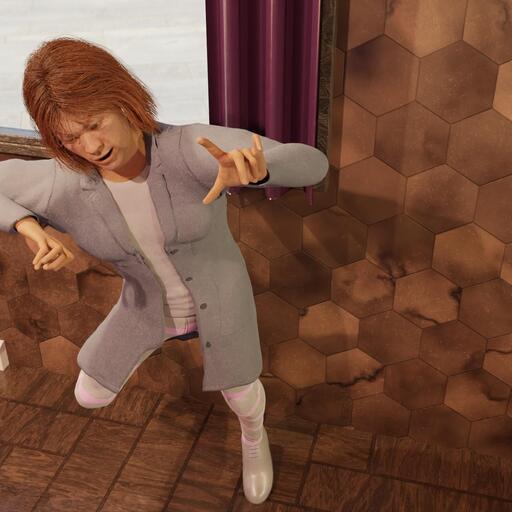}%
    \includegraphics[width=\annyonefigwidth]{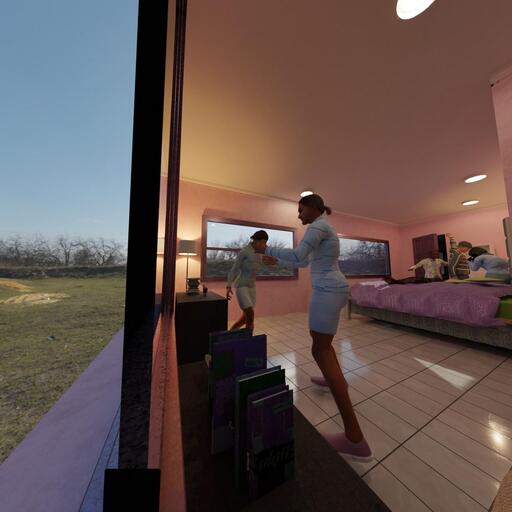}%
    \includegraphics[width=\annyonefigwidth]{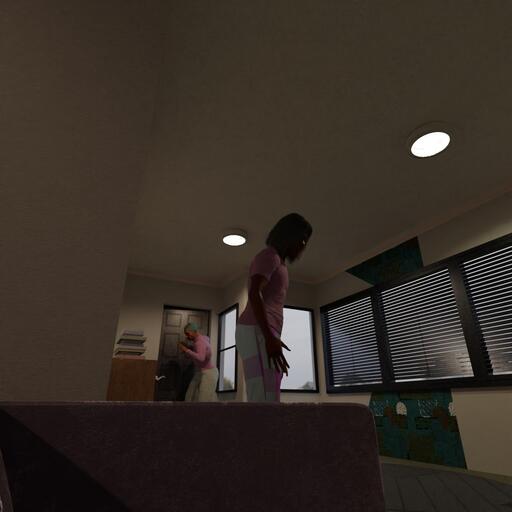}%
    \includegraphics[width=\annyonefigwidth]{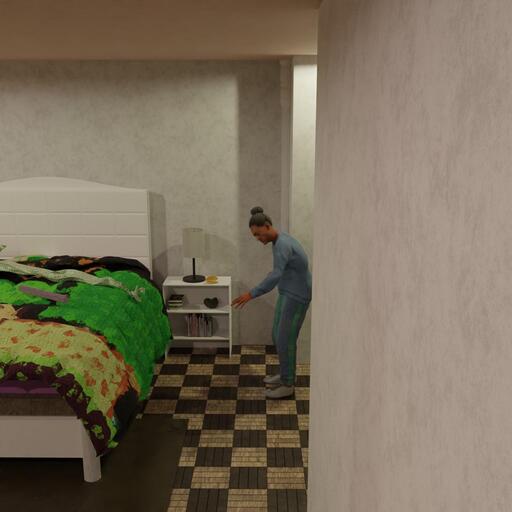}%
    
    \vspace{-1pt}
    \includegraphics[width=\annyonefigwidth]{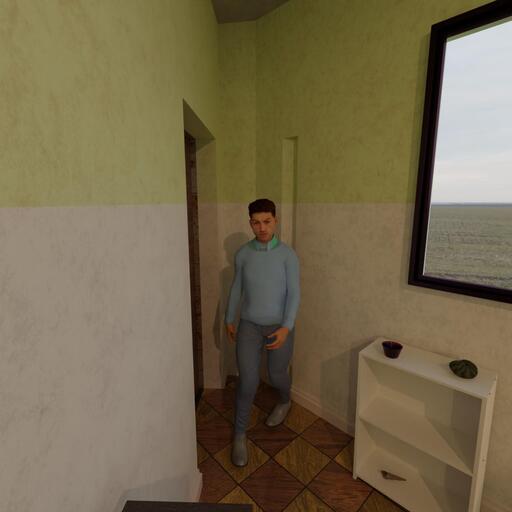}%
    \includegraphics[width=\annyonefigwidth]{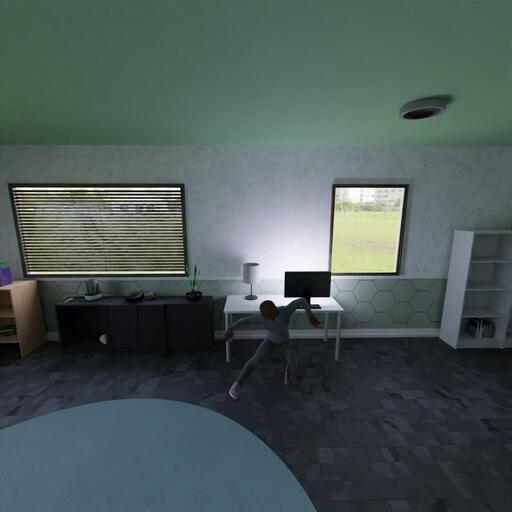}%
    \includegraphics[width=\annyonefigwidth]{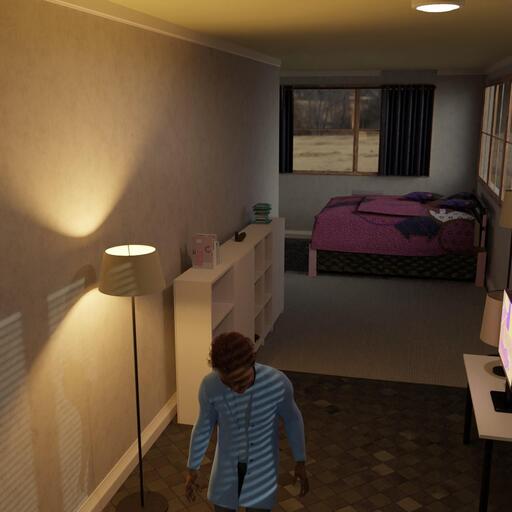}%
    \includegraphics[width=\annyonefigwidth]{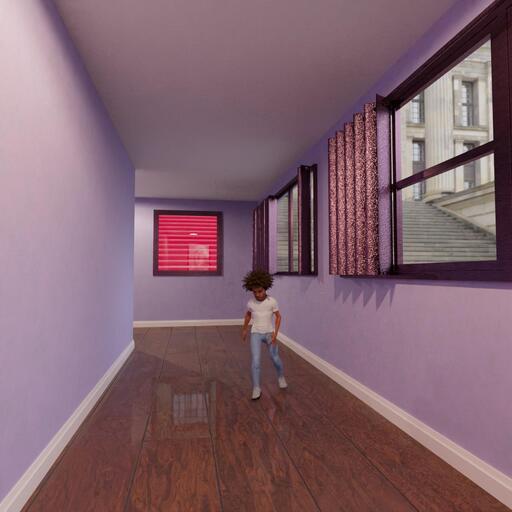}%
    \includegraphics[width=\annyonefigwidth]{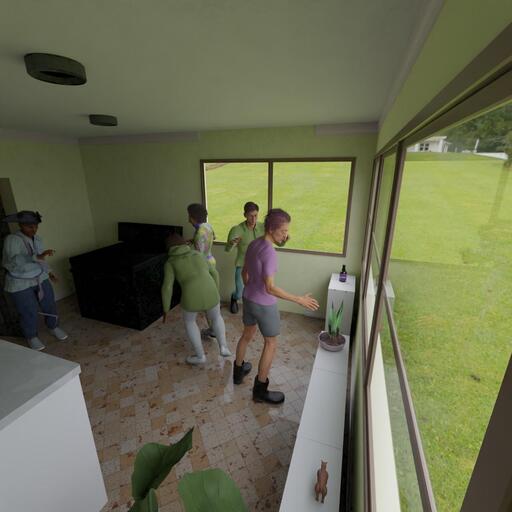}%
    \vspace{-0.2cm}
    \caption{\textbf{Samples from the \OurDataset dataset.}}
    \label{fig:dataset}
    \vspace{-0.2cm}
\end{figure}

For each individual in the dataset, we generate both a ground-truth annotation using Anny and a corresponding HumGen3D character.
Each character is randomly augmented with clothing and accessories sampled from the HumGen3D library, ensuring visual diversity.
These characters are then positioned within detailed, procedurally generated indoor scenes created using \textit{Infinigen Indoors}~\cite{infinigen2024indoors}.
Scene placement ensures no interpenetration between humans and surrounding objects.
On average, each scene includes approximately five people, and we render up to 40 camera views per scene, with camera placement biased toward human-centric framing.
We further enhance viewpoint diversity by rendering egocentric viewpoints, upper-body close-ups, and hand-focused views for a subset of the data.
The camera’s field of view is uniformly sampled between 30° and 130° to capture a wide range of spatial compositions.
Body poses are randomly sampled from AMASS~\cite{mahmood2019amass}, while hand poses are independently drawn from GRAB~\cite{taheri2020grab}.
Body shapes are sampled from a statistical distribution of human phenotypes derived from WHO population data (Section~\ref{sec:fitting}).
To ensure physical plausibility, we apply a self-collision check to eliminate invalid or unrealistic configurations.
All images are rendered using Blender Cycles~\cite{blender} at a resolution of 1280×1280 pixels.
Figure~\ref{fig:dataset} shows representative samples from \OurDataset, illustrating its diversity in body shapes, clothing, poses, and scene composition.
Overall, \OurDataset provides a large-scale, visually rich, and statistically diverse resource for training HMR models that transfer effectively to real-world data.

\subsection{HMR models}
\label{sub:hmrmodel}
To empirically evaluate our Anny body model and Anny-One dataset, we rely on two recent state-of-the-art HMR models: HMR2.0~\cite{goel2023humans} for the single-person setting and Multi-HMR~\cite{multi-hmr2024} for the multi-person setting. Both models are based on Vision transformers (ViTs). In both cases training code is available online, and we make minimal adaptations to predict Anny parameters instead of their original body models.

\paragraph{HMR2.0}~\cite{goel2023humans}
relies on the `Huge' variant of ViT with $16 \times 16$ patch size, followed by a transformer decoder that takes as input a single learned token that cross-attends to all image tokens to output body model parameters. It takes human-centric image crops as inputs.

\paragraph{Multi-HMR}~\cite{multi-hmr2024} is also built on a ViT backbone. It is pre-trained with DINOv2~\cite{dinov2} and released in various sizes (`Small', `Base' and `Large') with $14{\times}14$ patches. The ViT backbone is also followed by a cross-attention-based decoder that processes backbone output tokens corresponding to detected people. The model takes full, uncropped images as input, and is trained to detect humans, regress expressive human meshes, and place them in the scene in 3D.
We employ ViT-B ($448{\times}448$) for ablation studies and ViT-L ($672{\times}672$) for fair comparison with state-of-the-art methods.

\begin{table}[t]
\centering
\scriptsize
\caption{\textbf{Recovery of SMPL-X \vs Anny} body parameters on 3DPW and EHF. HMR2.0~\cite{goel2023humans} and Multi-HMR~\cite{multi-hmr2024} achieve comparable performances with both body models, under the same BEDLAM~\cite{bedlam} training setup.
}
\vspace{-0.2cm}
\begin{tabular}{lc|cc|cc}
 \multicolumn{2}{c|}{\textbf{Model}} & \multicolumn{2}{c|}{\textbf{3DPW}} & \multicolumn{2}{c}{\textbf{EHF}} \\
 Network & Body & MPJPE$\downarrow$ & PA-MPJPE$\downarrow$ & PVE$\downarrow$ & PA-PVE$\downarrow$ \\
\hline
\multirow{2}{*}{HMR2.0~\cite{goel2023humans}} & SMPL-X & \bf{86.0} & 52.0 & 76.4 & 66.9 \\
& Anny & 86.5 & \bf{49.4} & \bf{65.5} & \bf{49.7} \\
\midrule
\multirow{2}{*}{Multi-HMR~\cite{multi-hmr2024}} & SMPL-X & 87.1 & 56.3 & \bf{66.2} & 52.9 \\
& Anny & \bf{87.0} & \bf{54.3} & 68.6 & \bf{52.6} \\
\end{tabular}
\vspace{-0.2cm}
\label{tab:comparison}
\end{table}

\begin{table}[!b]
\centering
\vspace{-0.2cm}
\scriptsize
\setlength{\tabcolsep}{2pt}
\caption{\textbf{Impact of Body Model and Training Data} on AGORA~\cite{agora} validation set (featuring both adult and children). We use either BEDLAM or \OurDataset to pre-train a Multi-HMR~\cite{multi-hmr2024} model to regress SMPL-X/SMPL-X+A/Anny parameters.}
\vspace{-0.2cm}
\begin{tabular}{ccc|cc|cc}
 \multicolumn{3}{c|}{\textbf{Training}} & \multicolumn{2}{c|}{\textbf{AGORA-All}} & \multicolumn{2}{c}{\textbf{AGORA-Kids}} \\
 \small Pre-train & \small Fine-tune & \small Body & \small PVE$\downarrow$ & \small PA-PVE$\downarrow$ & \small PVE$\downarrow$ & \small PA-PVE$\downarrow$ \\
\hline
BEDLAM                  & \xmark & SMPL-X   & 140.8 & 71.3 & 186.0 & 64.2 \\
BEDLAM                  & \xmark & Anny     & 136.7 & 71.3 & 175.6 & 63.6 \\
\OurDataset             & \xmark & Anny     & 118.5 & 63.5 & 113.9 & 56.4 \\
\midrule
\xmark & \cmark & SMPL-X & 89.7 & 60.8 & 99.5  & 50.6 \\
\xmark & \cmark & SMPL-X+A & 87.9 & 60.4 & 80.8  & 50.5 \\
\xmark & \cmark & Anny   & 85.7 & 57.8 & 79.0  & 49.3 \\
\midrule
BEDLAM & \cmark & SMPL-X & 78.2 & 50.3 & 96.5 & 45.6 \\
BEDLAM & \cmark & SMPL-X+A & 76.6 & 50.0 & 77.6 & 43.6 \\
\OurDataset          & \cmark & Anny & \bf{72.8}  & \bf{48.2} & \bf{69.3} & \bf{41.5} \\

\end{tabular}
\vspace{-3mm}
\label{tab:shape}
\end{table}

\paragraph{Evaluation benchmarks.}
We evaluate 3D mesh prediction accuracy on standard benchmarks:  3DPW~\cite{vonMarcard2018}, EMDB~\cite{kaufmann2023emdb}, Hi4D~\cite{yin2023hi4d}, CMU-Toddler~\cite{cmu_panoptic_studio} and EHF~\cite{pavlakos2019smplx}.
We report the commonly used Mean Per Joint Position Error (MPJPE) and Per Vertex Error (PVE), along with their Procrustes-aligned variants, following prior work~\cite{conman,multi-hmr2024}.
We also report the Pair-PA-MPJPE, which measures the mean joint position error after Procrustes alignment of each pair of interacting humans, thereby evaluating the accuracy of their relative 3D poses.
Although the CMU-Toddler dataset contains both adults and children, its limited diversity among children prevents extensive ablation studies. To better assess performance across diverse populations, we use AGORA~\cite{agora}, which includes both adults and children. While AGORA is synthetic, it remains the only dataset currently available for this purpose.
Because the official validation set of AGORA does not contain any children, we re-define the training and validation sets such that images with human scans from 3DPEOPLE\cite{3dpeople} constitute the new validation set composed of 2k images, ensuring an equal percentage of children in both training and validation splits.

\subsection{HMR Results}

To isolate the effect of our body model from that of our synthetic dataset, we first evaluate Anny by re-training existing HMR methods on established datasets.
We then conduct experiments by training on Anny-One, optionally fine-tuning on the training set of the respective benchmarks.
Finally we train using both \OurDataset and standards HMR training data~\cite{kolotouros2019learning} including BEDLAM~\cite{bedlam}, MS-COCO~\cite{lin2014microsoft}, and MPII~\cite{andriluka14cvpr} to compare against state-of-the-art methods.

\begin{table*}[!t]
\centering
\scriptsize
\setlength{\tabcolsep}{3pt} %
\caption{
\textbf{Comparison to state-of-the-art image-based methods.}
Quantitative results on multiple datasets comparing our approach against existing multi-person image-based methods.
Lower values indicate better performance~($\downarrow$).
}
\vspace{-2mm}
\newcommand{\mylegend}[1]{\rotatebox{90}{\tiny #1}}
\scriptsize
\begin{tabular}{l|ccc|ccc|ccc|cc}
& \multicolumn{3}{c|}{\textbf{3DPW}} 
& \multicolumn{3}{c|}{\textbf{EMDB}} 
& \multicolumn{3}{c|}{\textbf{Hi4D}}
& \multicolumn{2}{c}{\textbf{CMU-Toddler}} \\
\textbf{Model} & \mylegend{PA-MPJPE} & \mylegend{MPJPE} & \mylegend{PVE}
 & \mylegend{PA-MPJPE} & \mylegend{MPJPE} & \mylegend{PVE}
 & \mylegend{PA-MPJPE} & \mylegend{MPJPE} & \mylegend{Pair-PA-MPJPE}
 & \mylegend{MPJPE} & \mylegend{Pair-PA-MPJPE} \\
\hline
AiOS~\cite{sun2024aios} & 45.0 & 68.8 & 90.9 & 63.3 & 90.6 & 108.1 & 49.9 & 71.4 & 234.2 & 162.4 & 723.1 \\
SAT-HMR~\cite{su2025sat} & 52.7 & 81.0 & 94.5 & 71.0 & 112.9 & 126.7 & 61.2 & 88.2 & 85.5 & 153.9 & 654.8\\
Multi-HMR~\cite{multi-hmr2024} &  46.9 & \bf{69.5} & 88.8 & 48.5 & 73.7 & 87.1 & 49.8 & 67.8 & 80.6 & 153.6 & 638.9 \\
\bf{Multi-HMR+Anny} & 
\bf{41.8} & 71.5 & \bf{83.2} & 
\bf{48.5} & \bf{71.5} & \bf{83.4} & 
\bf{48.7} & \bf{66.6} & \bf{80.0} & 
\bf{102.1} & \bf{263.8}  \\ %
\end{tabular}
\label{tab:sota}
\end{table*}

\begin{figure}[!t]
	\centering

	\begin{subfigure}{0.56\linewidth}
		\centering
		\tiny
		\includegraphics[width=0.666\linewidth,clip,trim=0px 500px 0px 300px]{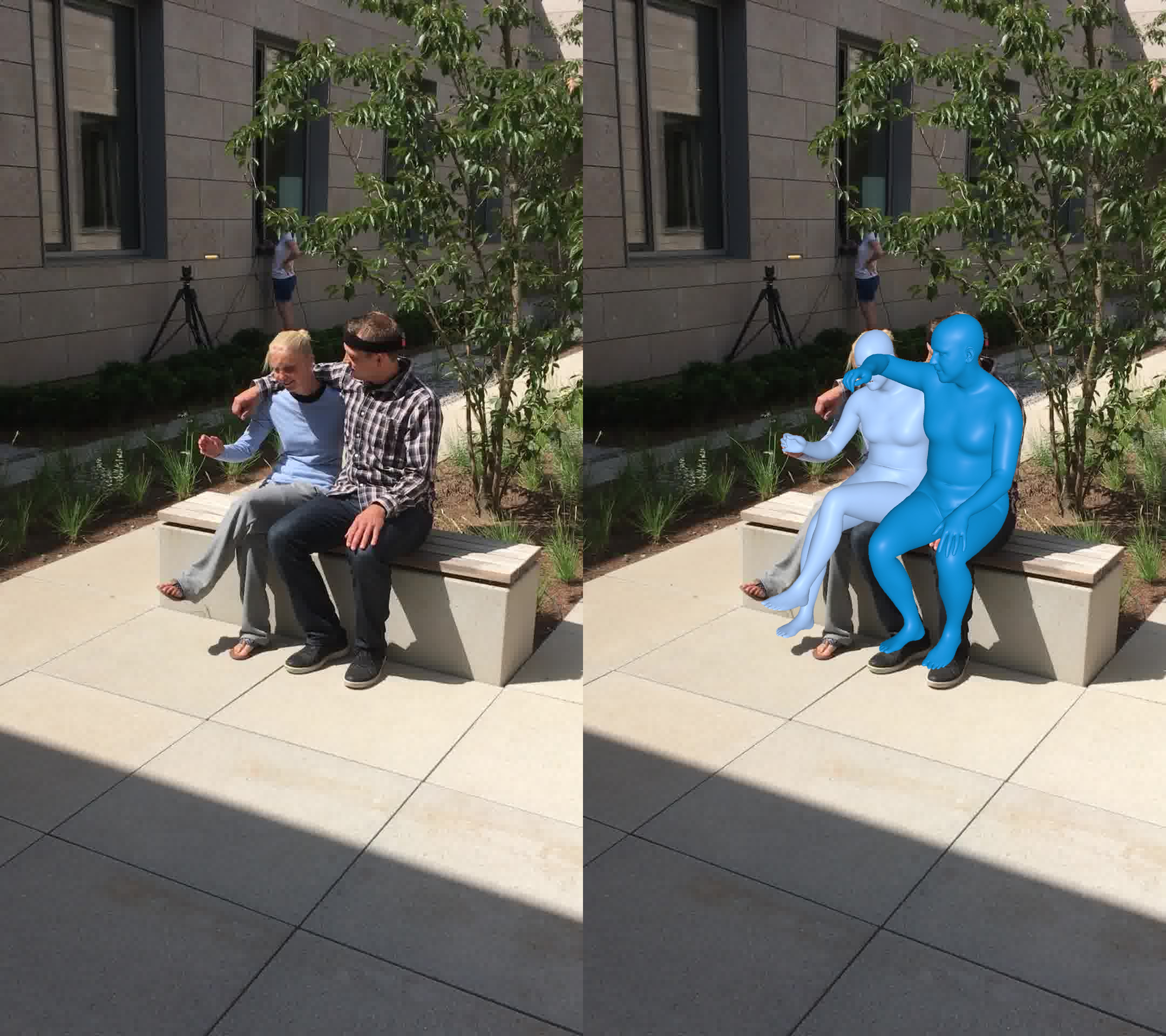}%
		\includegraphics[width=0.332\linewidth,clip,trim=1080px 500px 0px 300px]{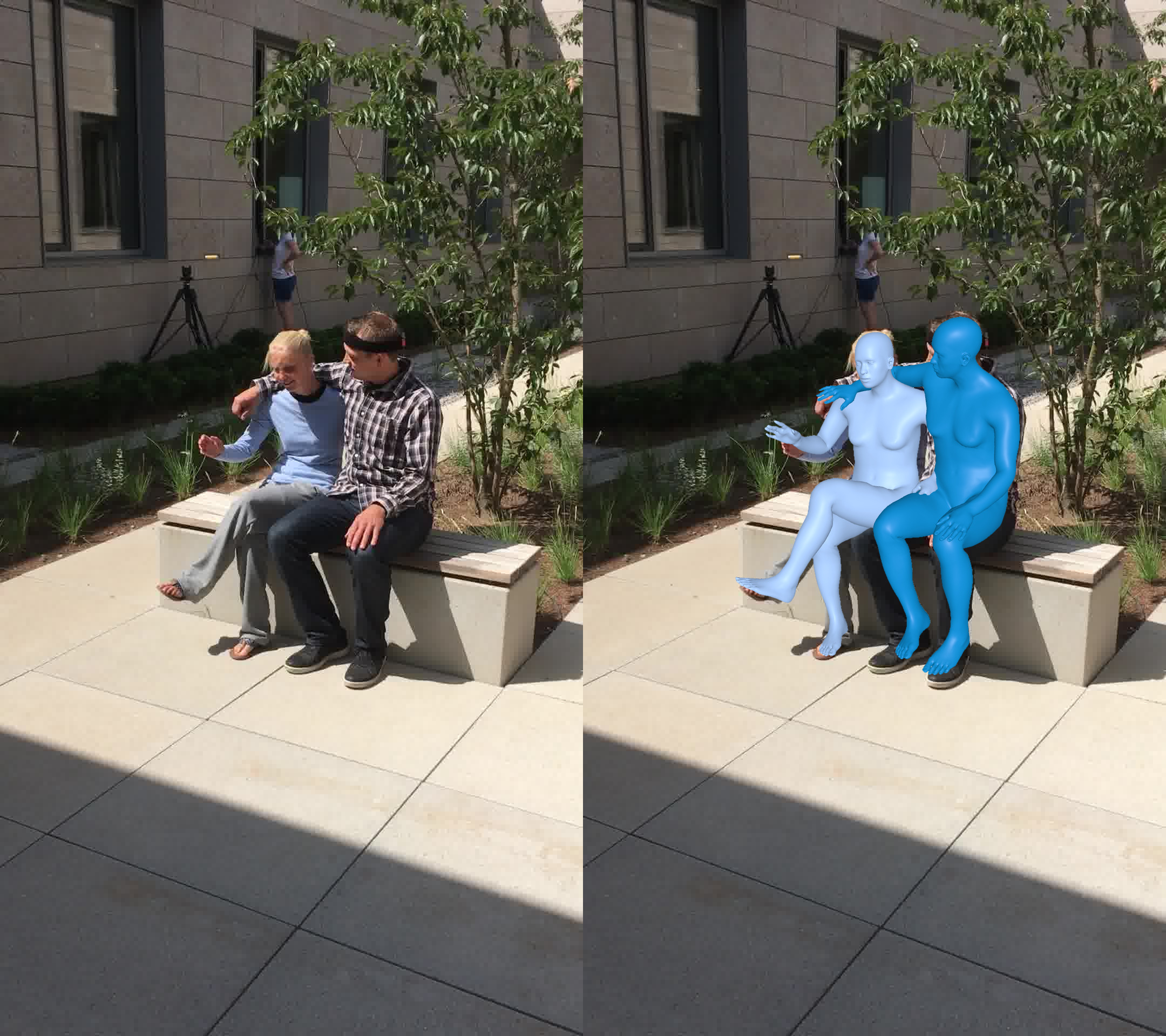}
		
	\end{subfigure}
	\hfill
	\begin{subfigure}{0.42\linewidth}
		\centering
		\tiny
		\includegraphics[width=0.666\linewidth,clip,trim=0px 150px 0px 300px]{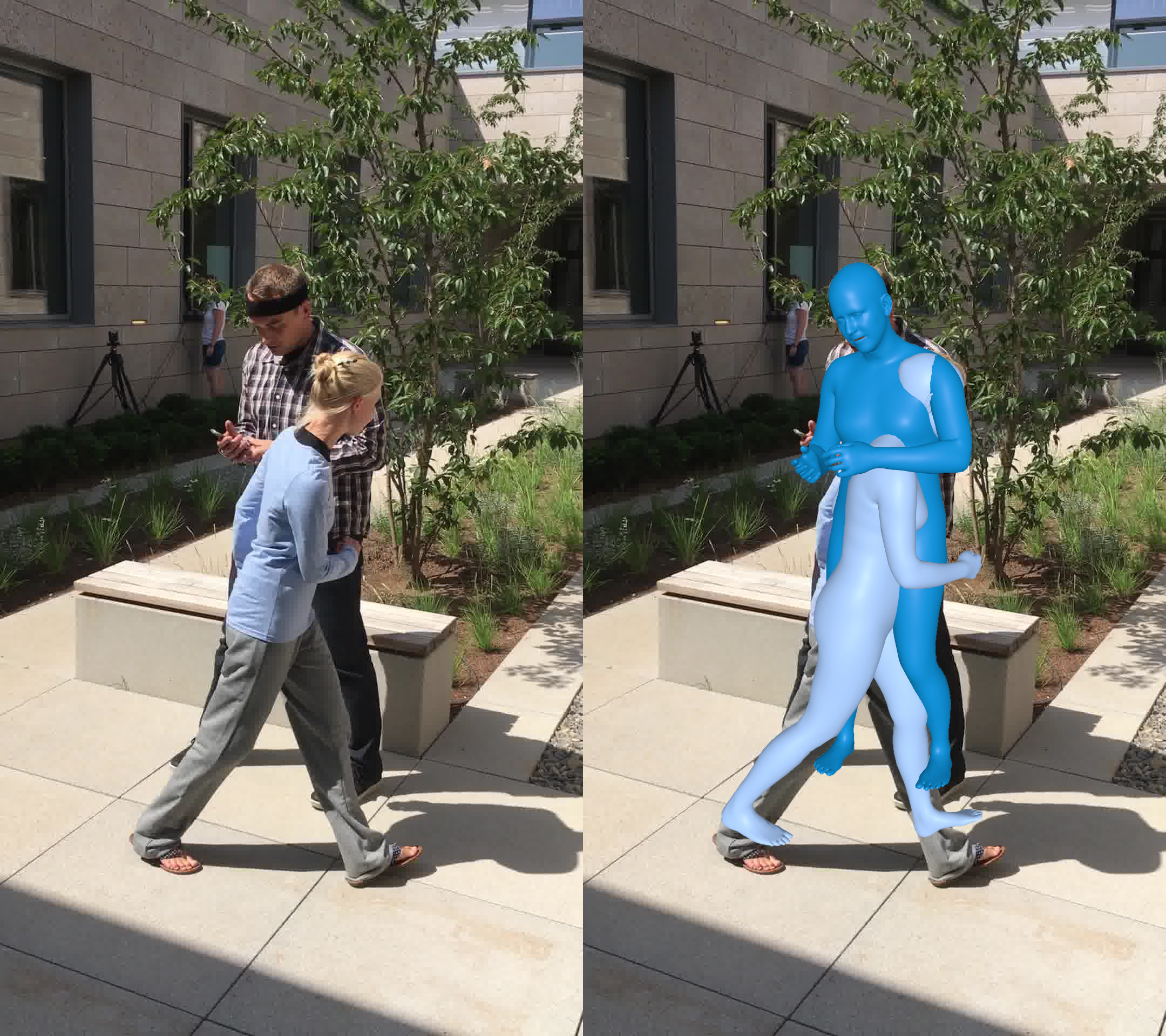}%
		\includegraphics[width=0.332\linewidth,clip,trim=1080px 150px 0px 300px]{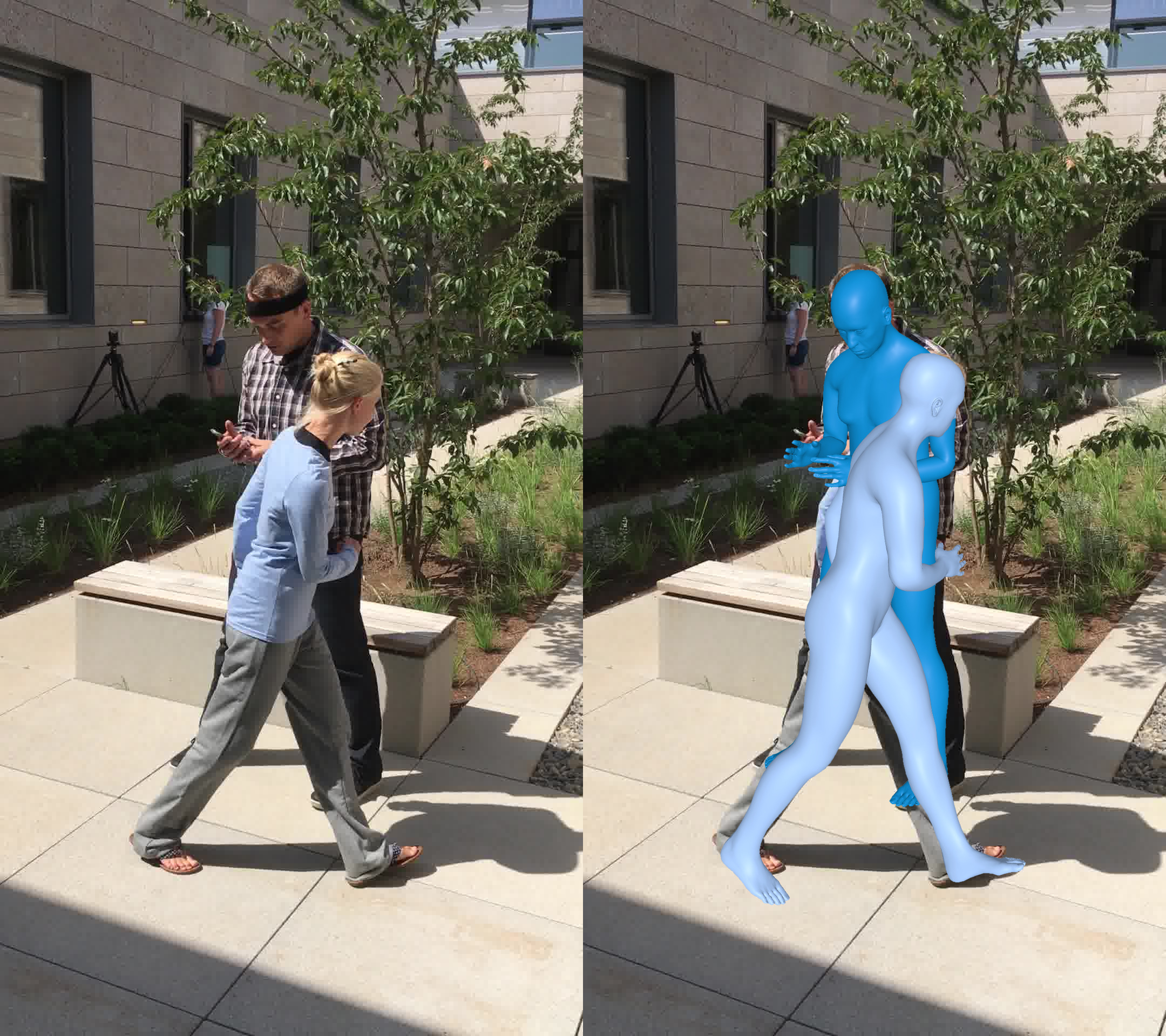}
		
	\end{subfigure}
	
	\vspace{0.01mm}
	
	\begin{subfigure}{0.56\linewidth}
		\centering
		\tiny
		\includegraphics[width=0.666\linewidth,clip,trim=0px 100px 0px 200px]{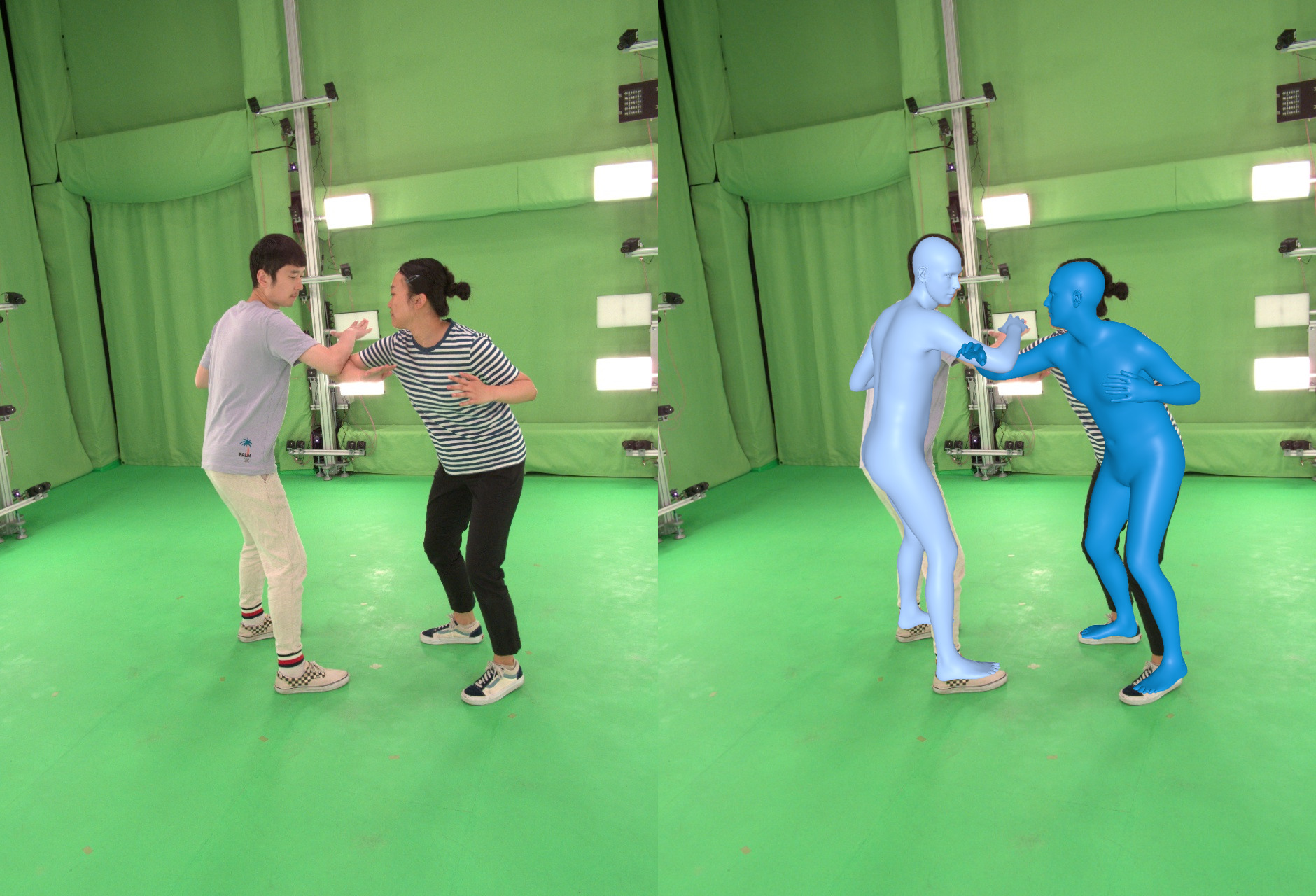}%
		\includegraphics[width=0.332\linewidth,clip,trim=945px 100px 0px 200px]{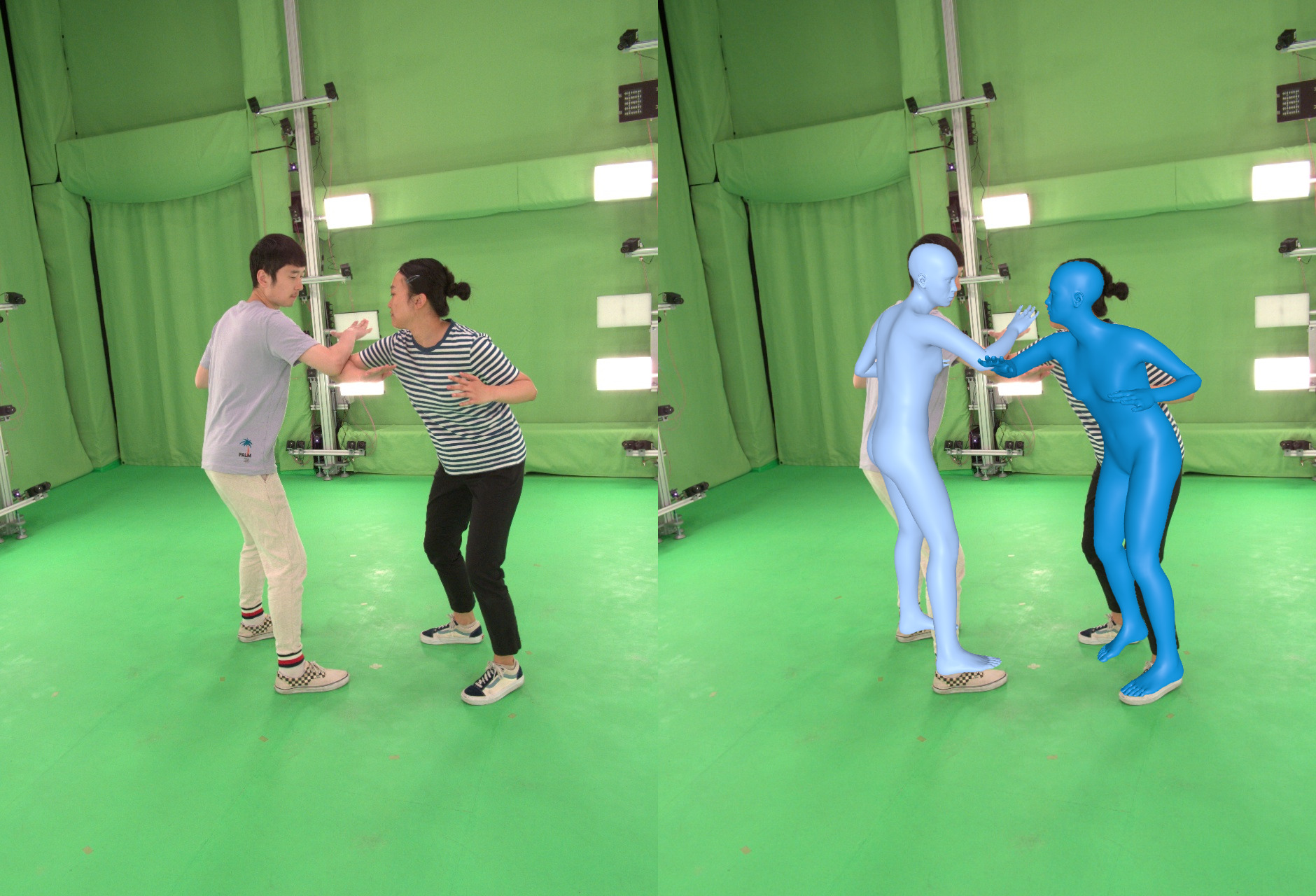}
		
		\begin{minipage}[c]{0.32\linewidth}
			\centering
			Input
		\end{minipage}
		\begin{minipage}[c]{0.32\linewidth}
			\centering
			SMPL-X
		\end{minipage}
		\begin{minipage}[c]{0.32\linewidth}
			\centering
			Anny
		\end{minipage}
	\end{subfigure}
	\hfill
	\begin{subfigure}{0.42\linewidth}
		\centering
		\tiny
		\includegraphics[width=0.666\linewidth,clip,trim=0px 0px 0px 0px]{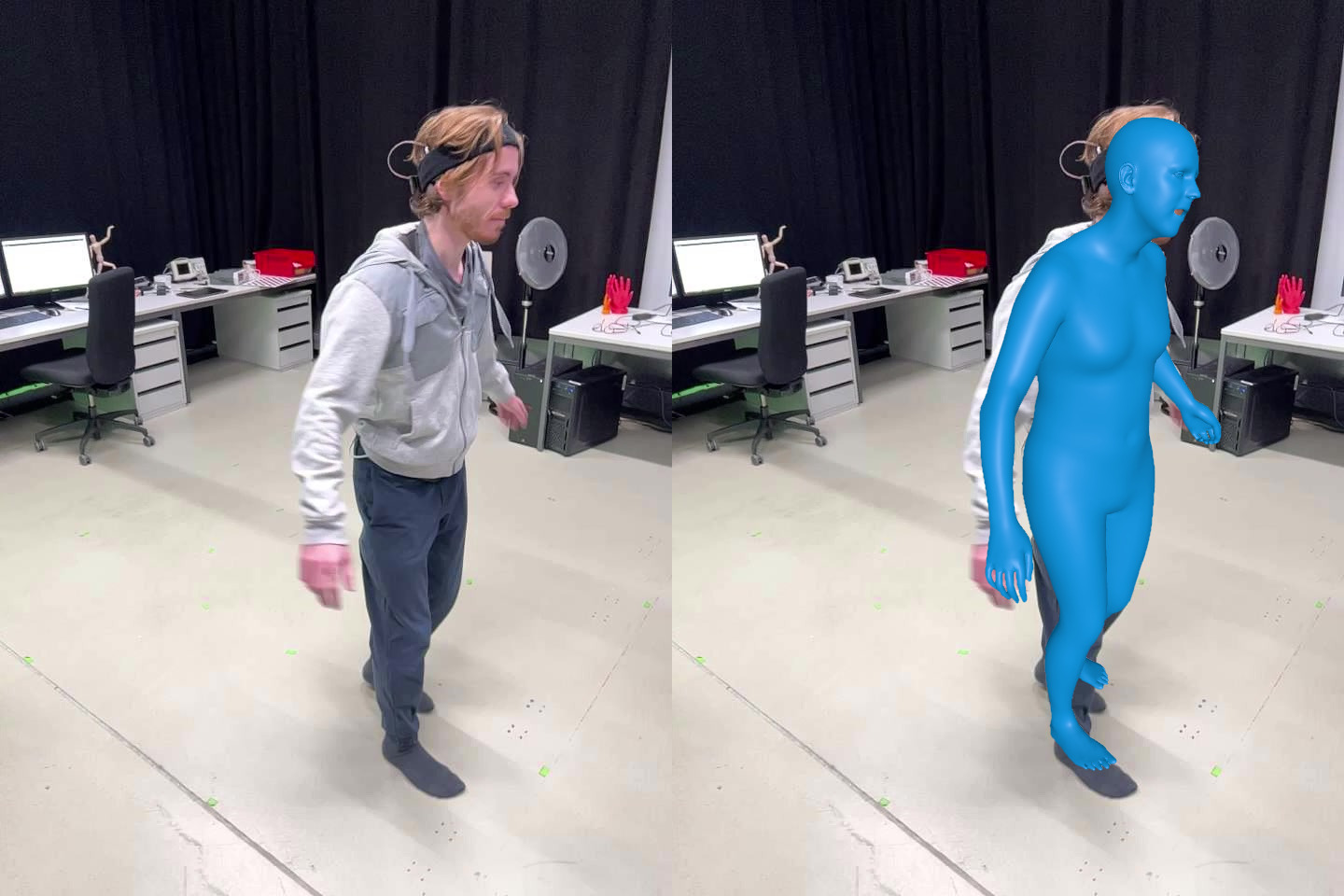}%
		\includegraphics[width=0.332\linewidth,clip,trim=724px 0px 0px 0px]{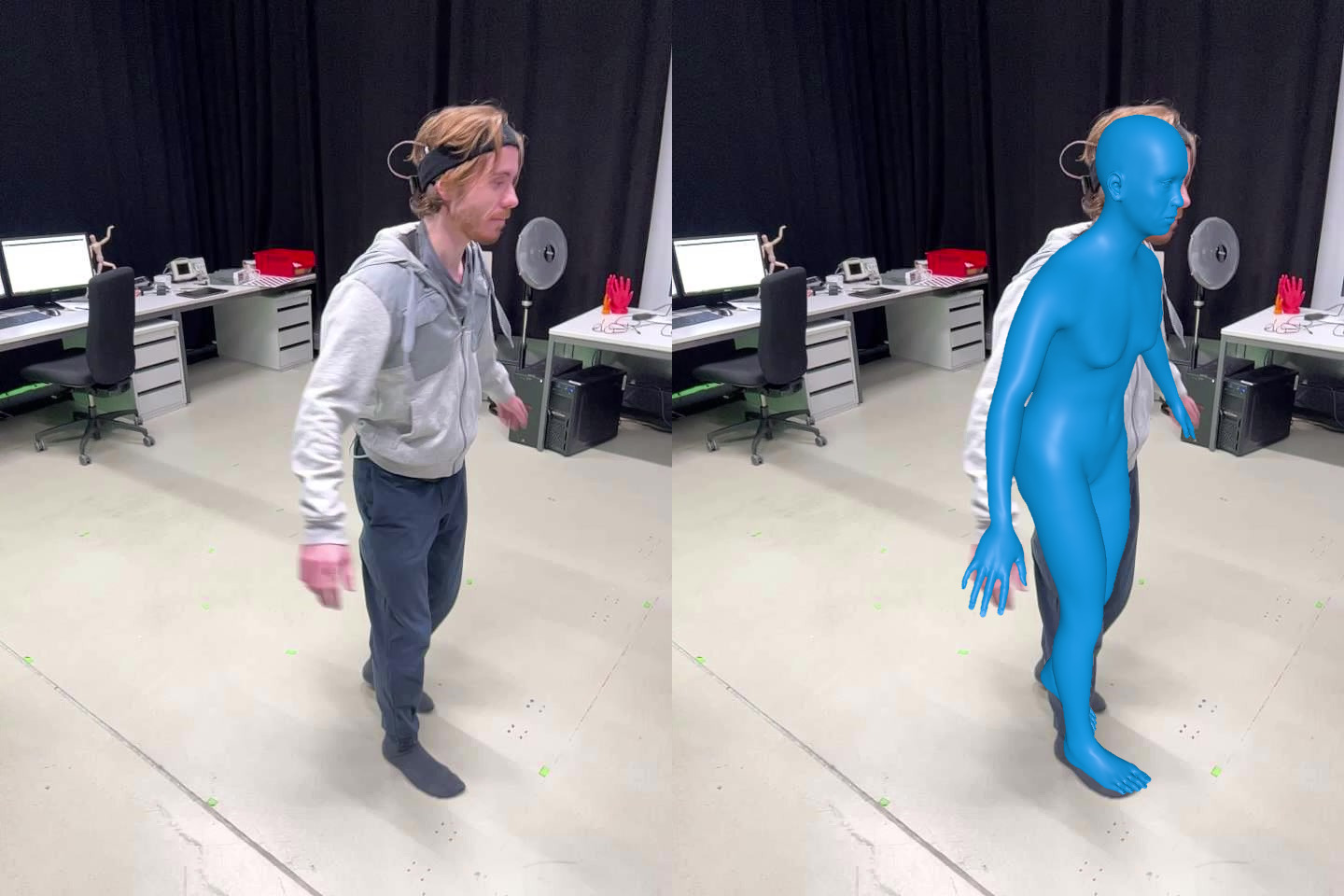}
		
		\begin{minipage}[c]{0.32\linewidth}
			\centering
			Input
		\end{minipage}
		\begin{minipage}[c]{0.32\linewidth}
			\centering
			SMPL-X
		\end{minipage}
		\begin{minipage}[c]{0.32\linewidth}
			\centering
			Anny
		\end{minipage}
	\end{subfigure}
	
	\vspace{-3mm}
	\caption{\label{fig:multihmr_anny_smplx_comparison}
		\textbf{Qualitative results} comparison between Multi-HMR trained with SMPL-X and with Anny on academic benchmarks: 3DPW (top), Hi4D (bottom left), and EMDB (bottom right).
	}
\end{figure}

\paragraph{\ours \vs SMPL-X: a scan-free body model is sufficient.}
We adapt the implementations of two foundation models (HMR2.0~\cite{goel2023humans} and Multi-HMR~\cite{multi-hmr2024}) to predict full-body meshes using either \ours or the SMPL-X~\cite{pavlakos2019smplx} body model.
We train the HMR methods on BEDLAM and evaluate them on the 3DPW and EHF test sets.
To limit the computational cost of experiments, HMR2.0 is finetuned with a frozen backbone and without data augmentation. We also restrict the modeling of Anny shapes to 6 main phenotype parameters (age, gender, height, weight, muscle, proportions).
Results are reported in Table~\ref{tab:comparison}.
Across all metrics, using \ours achieves comparable or superior performance to SMPL-X, despite its non data-driven design.
This demonstrates that a %
body model such as \ours is sufficient for HMR and can serve as a drop-in replacement for SMPL-X.
The gains observed with Anny may be attributable to its particular shape and pose parameterization; however its precise influence on machine learning performance remains unclear. By making Anny publicly available, we aim to facilitate further analysis and encourage deeper investigation.

\paragraph{Anny and \OurDataset: modeling diverse body shapes.}
We evaluate performance on AGORA, which is the only existing standard benchmark containing children, and report results in Table~\ref{tab:shape}. 
First we train models on BEDLAM with SMPL-X or Anny. We see in the first and second rows of Table~\ref{tab:shape} that using the Anny head marginally improves performance, which may be due to the fact that Anny handles children more gracefully.
Our proposed dataset consists of $780$k synthetic images, designed to contain humans with diverse body shapes, diverse backgrounds and changes in camera intrinsics.
Training on Anny-One instead of BEDLAM (third row) brings substantial gains, in particular for children.  

We then consider training directly on AGORA, and compare Anny to both SMPL-X+A and SMPL-X. Both SMPL-X+A and Anny perform significantly better than SMPL-X, with a moderate advantage for Anny. This is consistent with the fact that they were both designed to handle children.
Finally, we consider the full-data regime with pretraining on either BEDLAM or Anny-One, followed by finetuning on AGORA (last three rows of the table). We observe that in this large-scale regime, the combination of Anny-One and Anny significantly outperforms existing datasets and body models on AGORA. 

\paragraph{Comparison to state-of-the-art methods.}
We evaluate Multi-HMR trained with \ours on a large-scale mixture of datasets, including \OurDataset and standard HMR training data (BEDLAM, MS-COCO, MPII), and compare against state-of-the-art multi-person HMR approaches.
Results on 3DPW, EMDB, Hi4D, and CMU-Toddler (Table~\ref{tab:sota} and Figure~\ref{fig:multihmr_anny_smplx_comparison})
show that our model achieves competitive performances 
across all benchmarks.

These results demonstrate that \ours (i.e. Multi-HMR with \ours) scales effectively to large and diverse data, enabling robust reconstruction of adults and children alike across both in-the-wild and controlled multi-person scenarios.
In particular, the results on CMU-Toddler suggest that Anny is able to model a broad range of body shapes, including both adults and children.
Overall, these results demonstrate that a scan-free body model such as Anny can achieve performance comparable to scan-based models like SMPL-X, while providing a simple, unified, and interpretable representation for modern HMR tasks.

\begin{figure*}[t]
	\setlength{\tabcolsep}{1pt}
	\renewcommand{\arraystretch}{0.}
	\begin{tabular}{ccc}
		\includegraphics[width=0.33\linewidth, trim=0cm 5cm 0cm 5cm, clip]{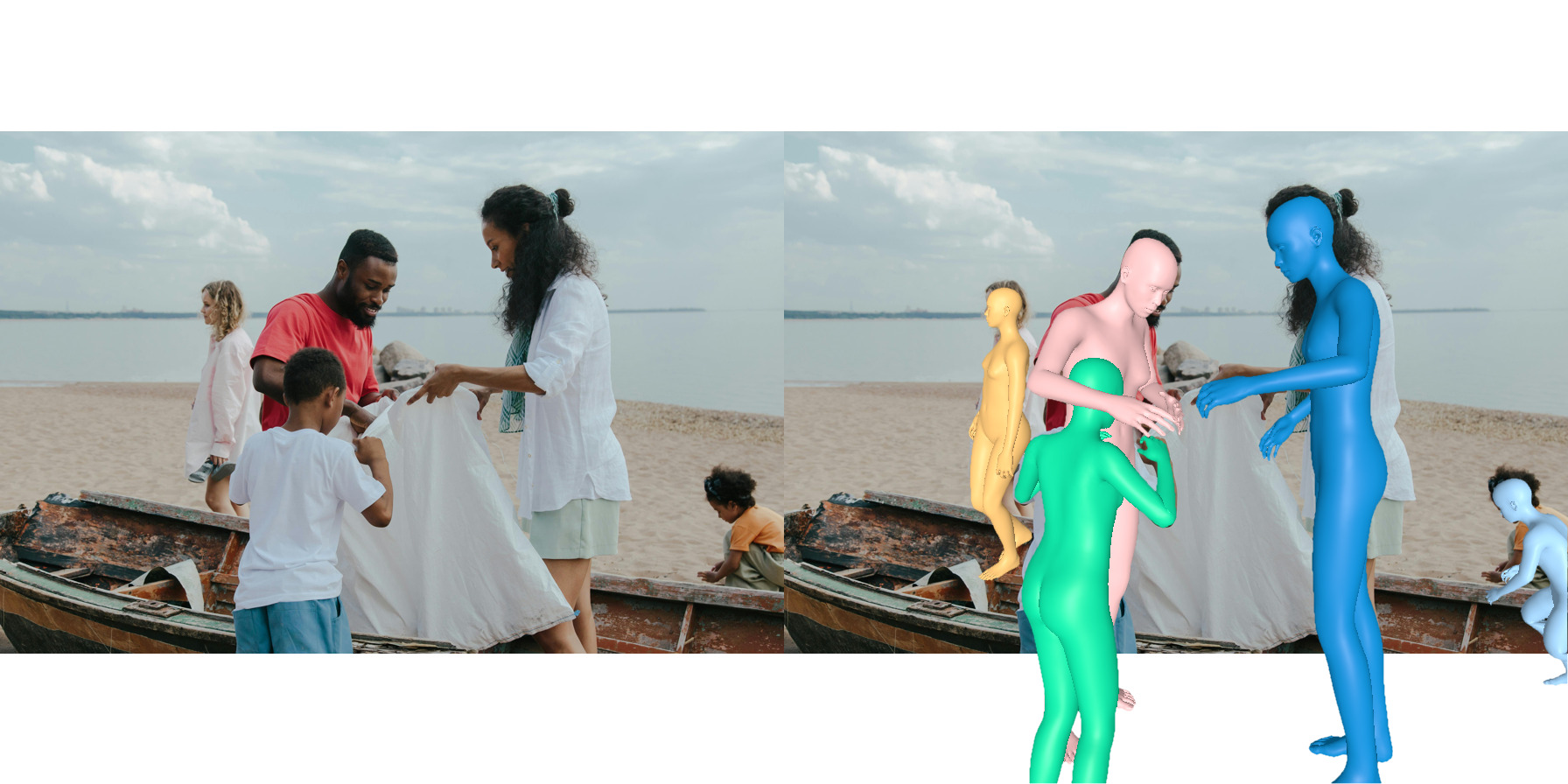}& 
		\includegraphics[width=0.33\linewidth, trim=0cm 5cm 0cm 5cm, clip]{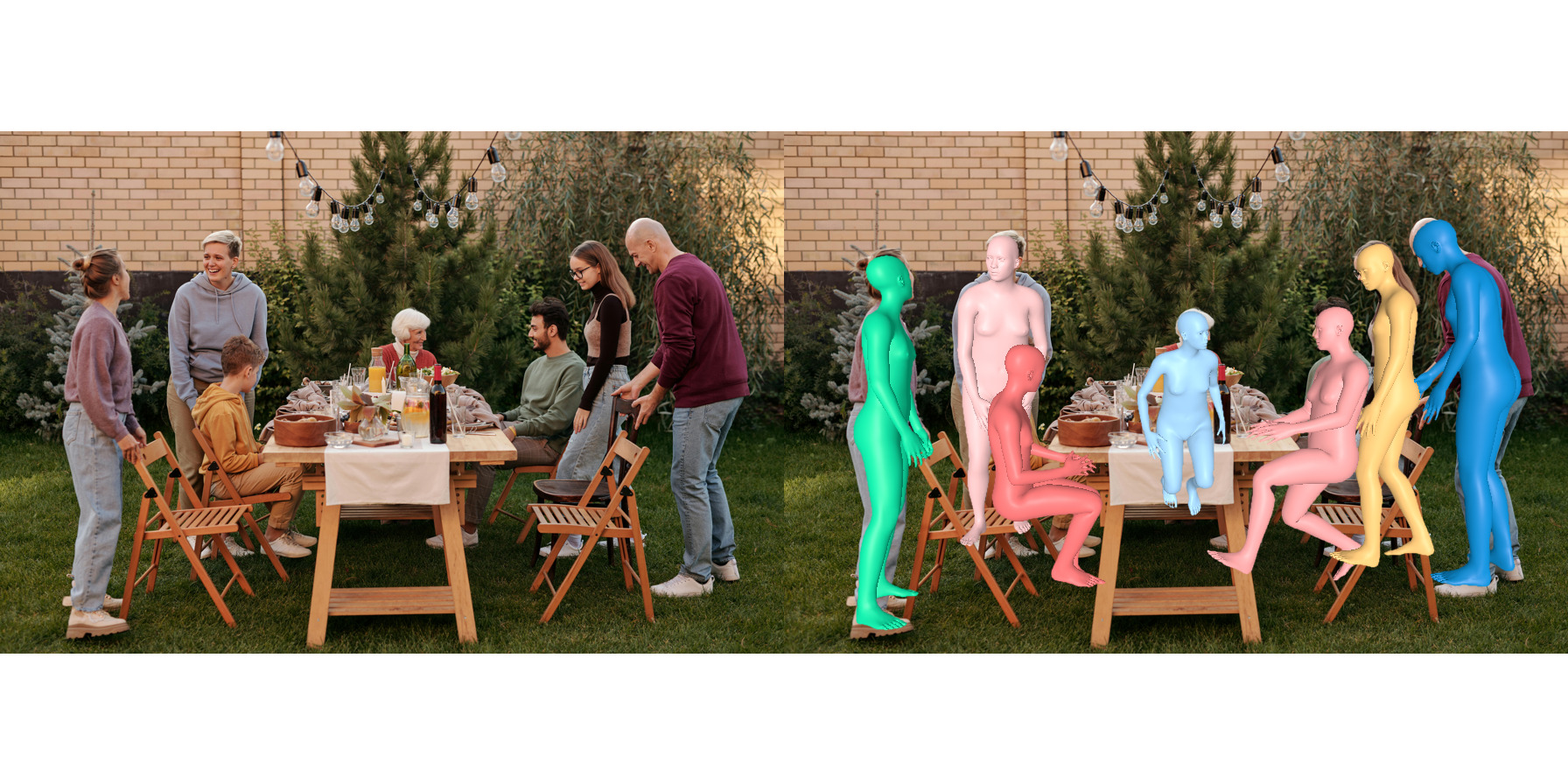}& 
		\includegraphics[width=0.33\linewidth, trim=0cm 5cm 0cm 5cm, clip]{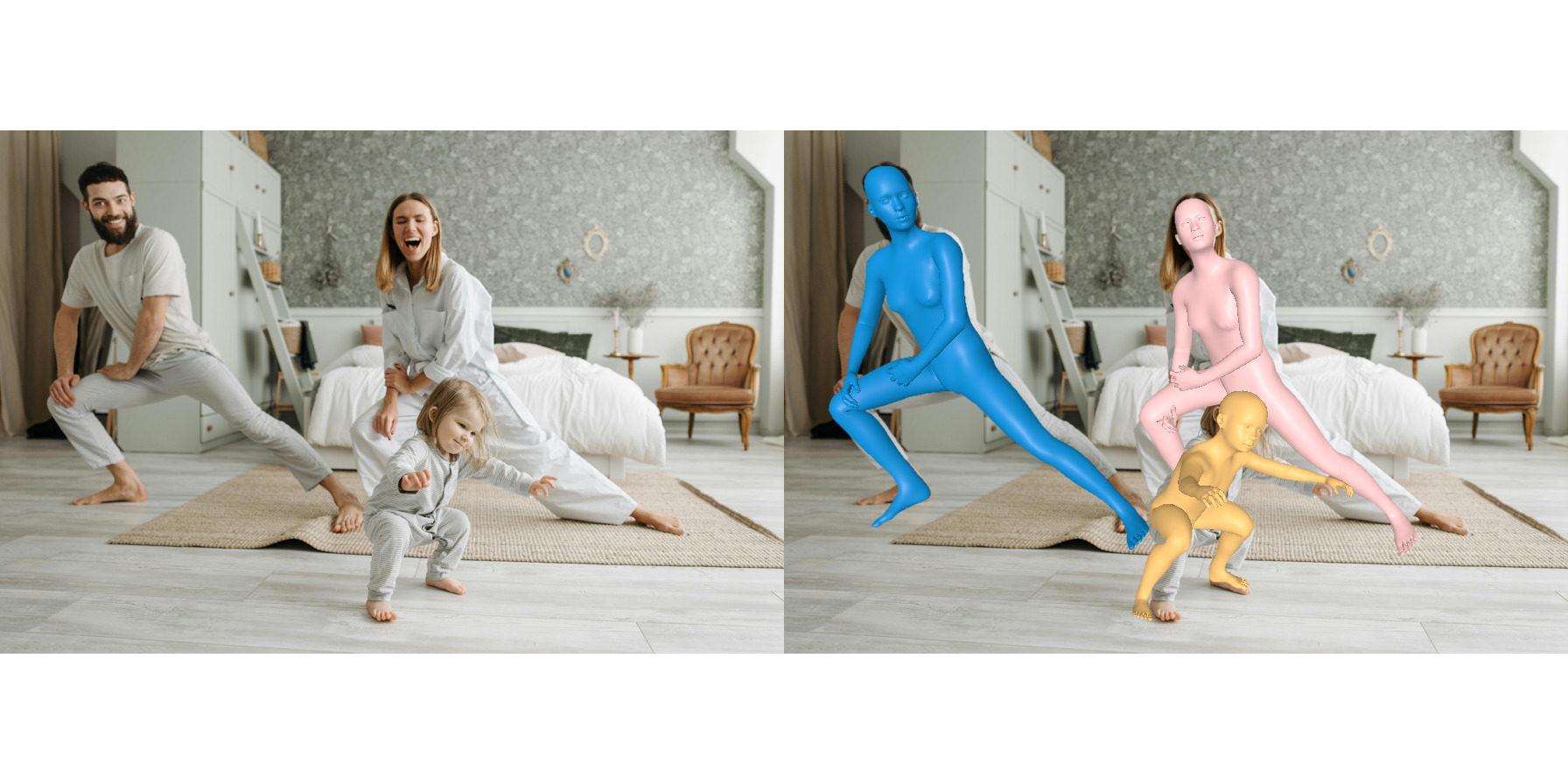}\\[0cm]
		\includegraphics[width=0.33\linewidth, trim=0cm 5cm 0cm 5cm, clip]{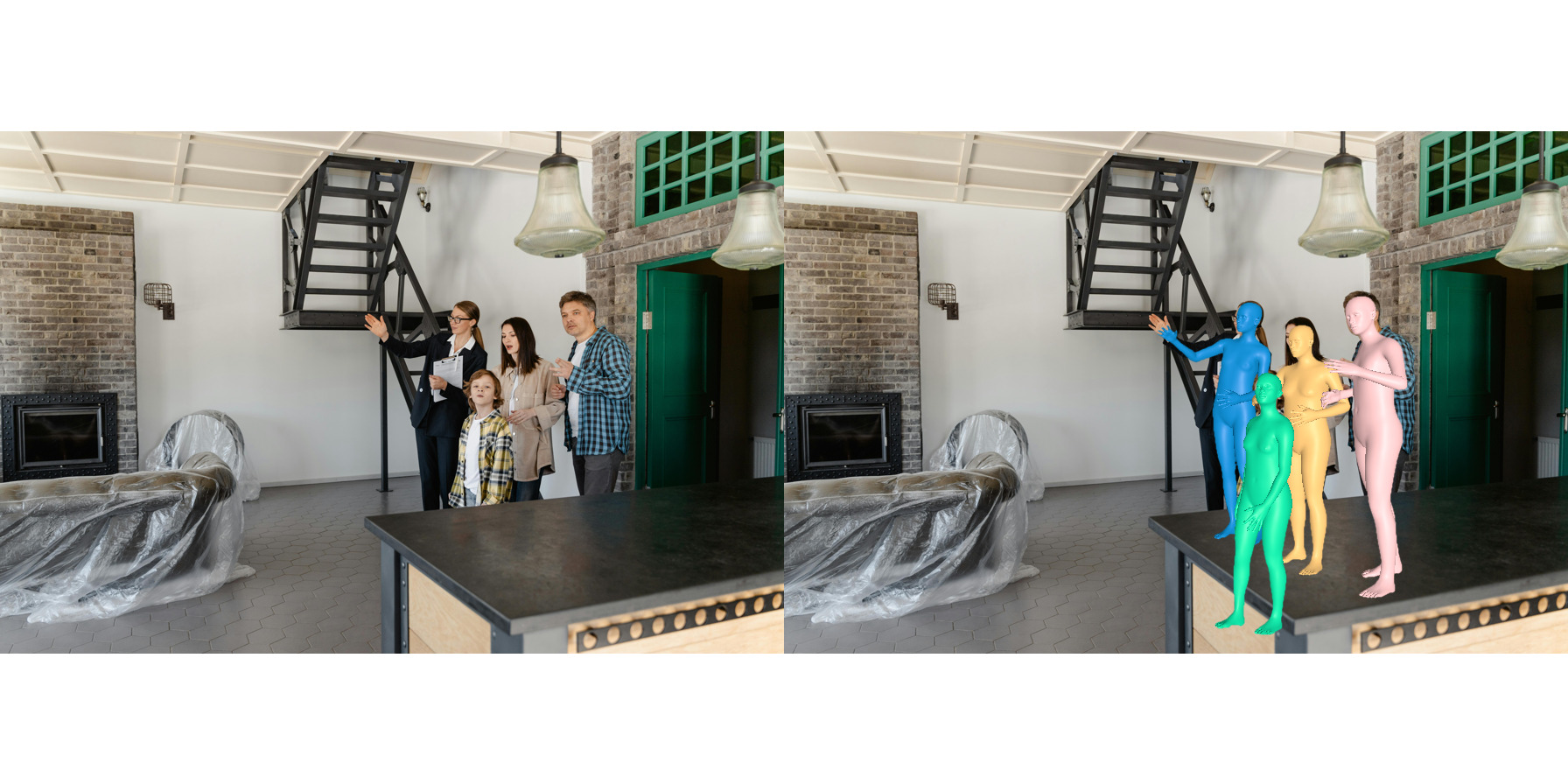}& 
		\includegraphics[width=0.33\linewidth, trim=0cm 5cm 0cm 5cm, clip]{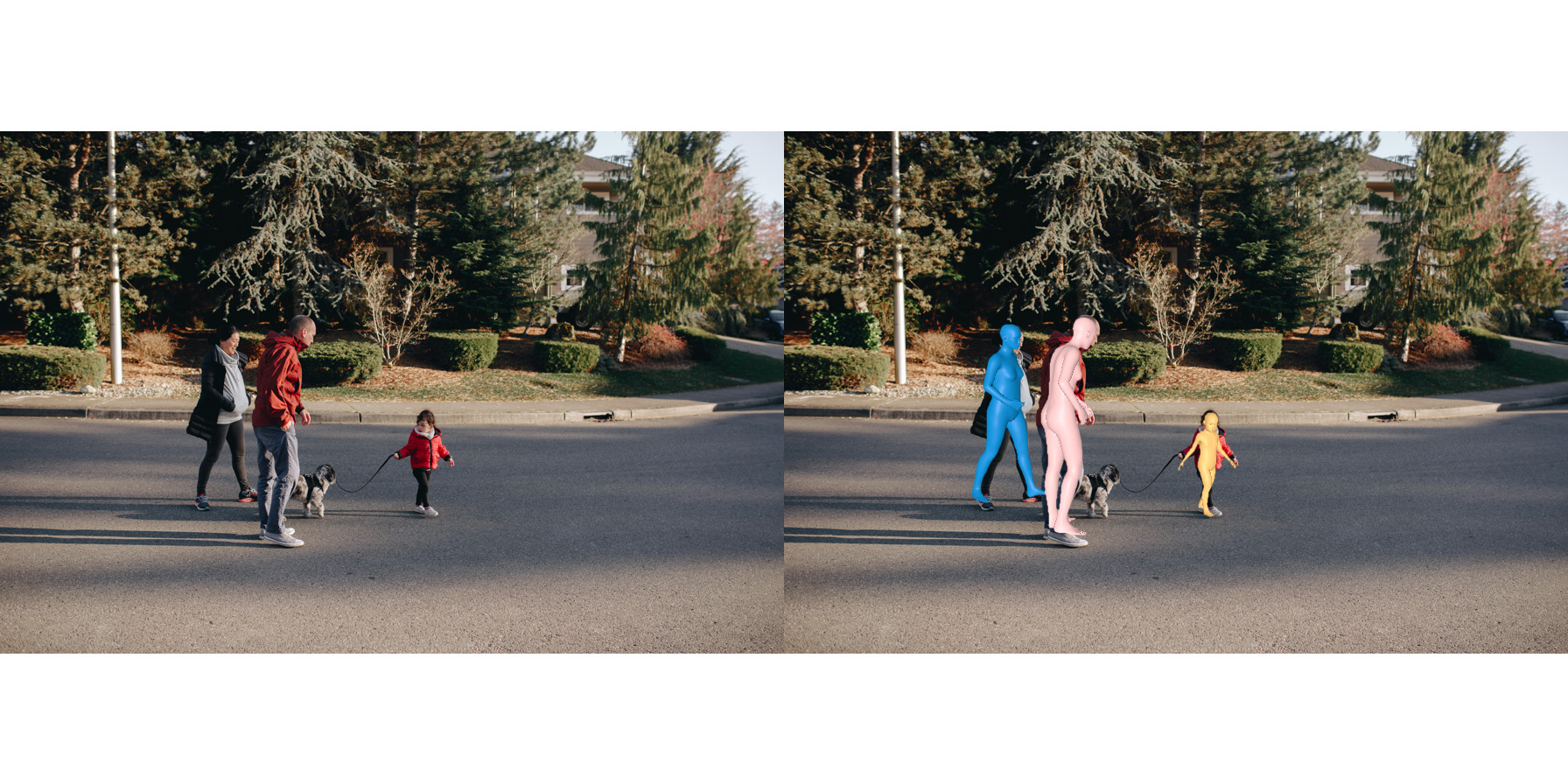}&
		\includegraphics[width=0.33\linewidth, trim=0cm 5cm 0cm 5cm, clip]{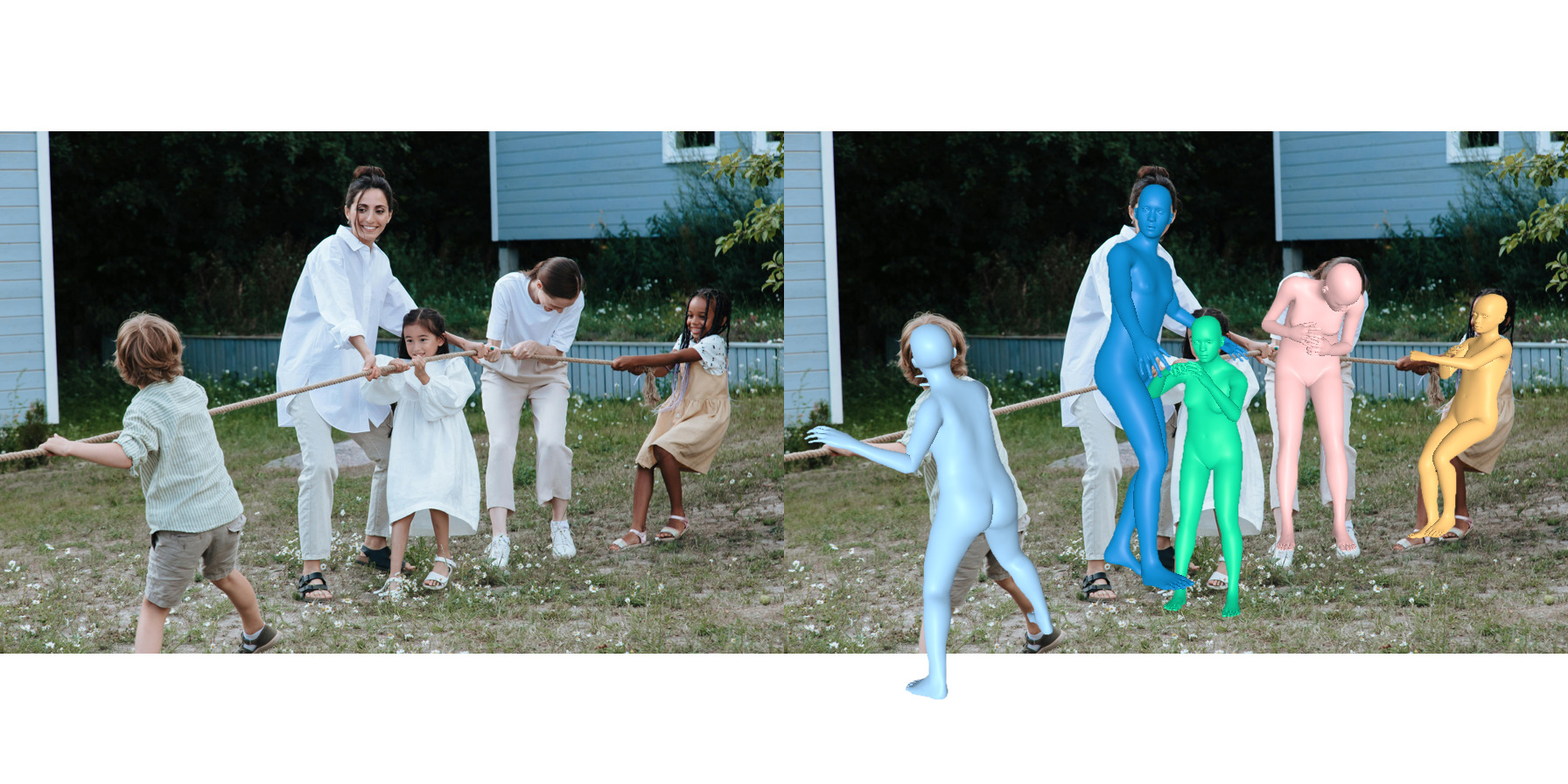}\\[0cm]
		\includegraphics[width=0.33\linewidth, trim=0cm 5cm 0cm 5cm, clip]{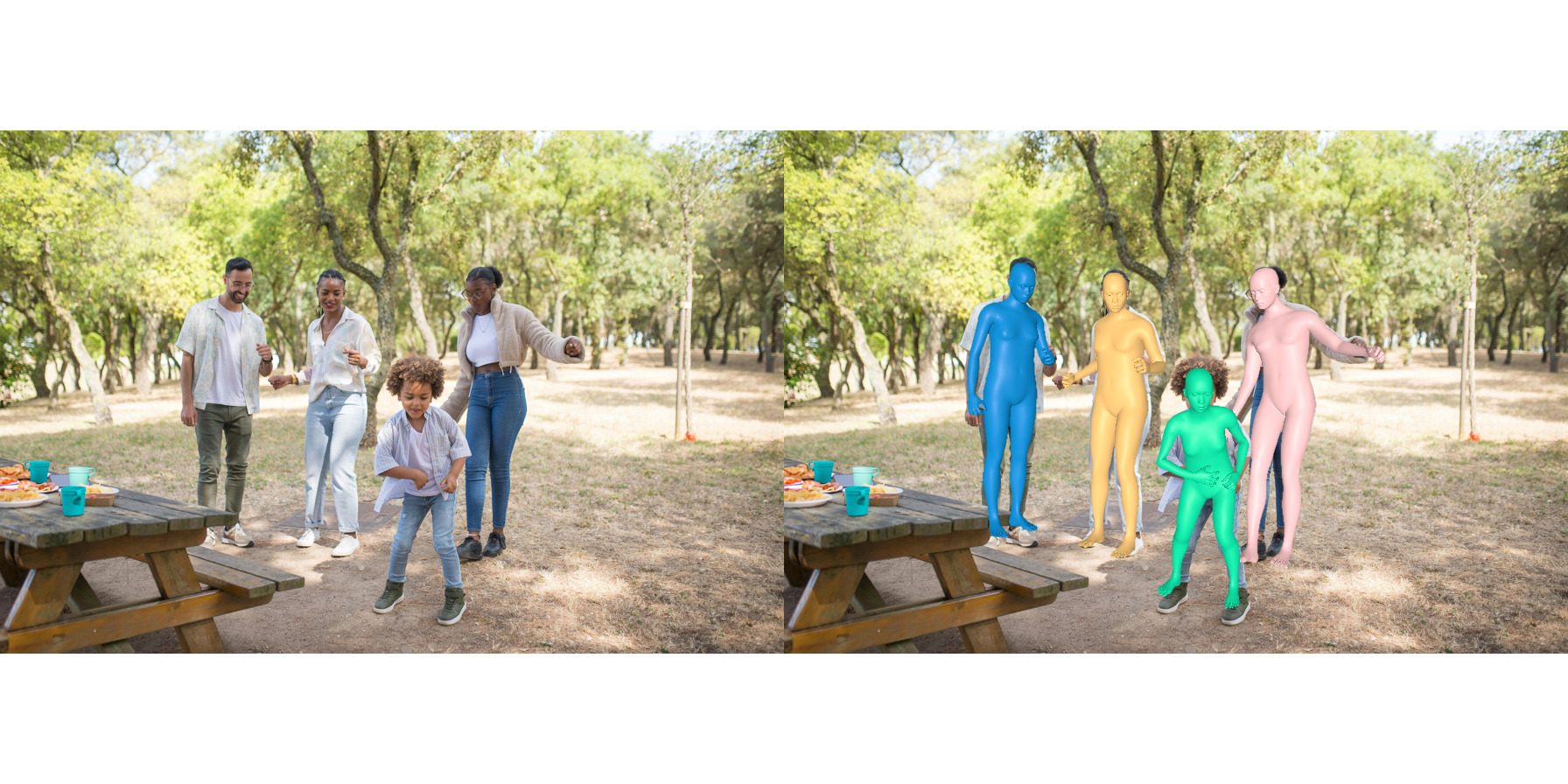}&
		\includegraphics[width=0.33\linewidth, trim=0cm 5cm 0cm 5cm, clip]{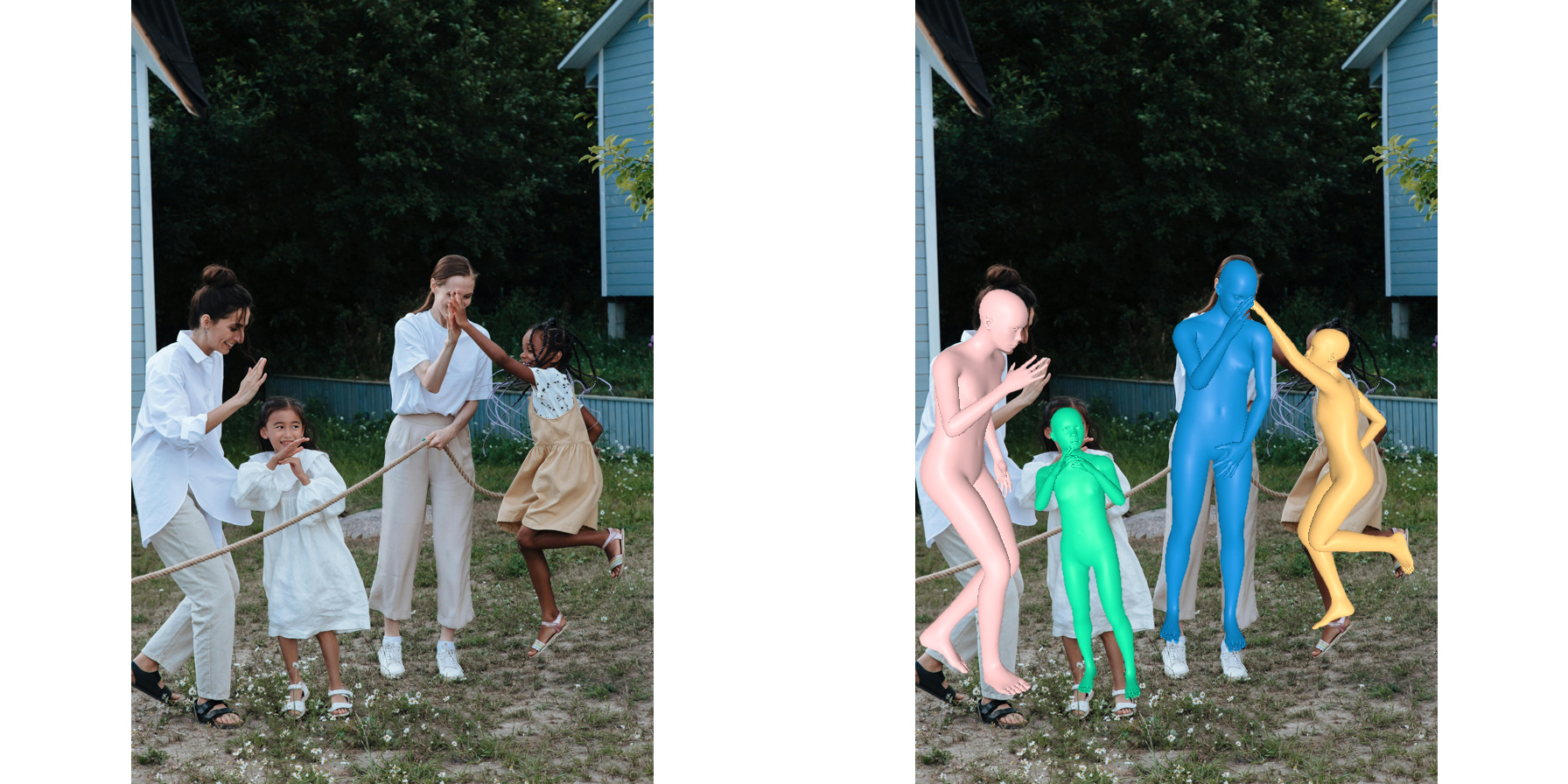}&
		\includegraphics[width=0.33\linewidth, trim=0cm 5cm 0cm 5cm, clip]{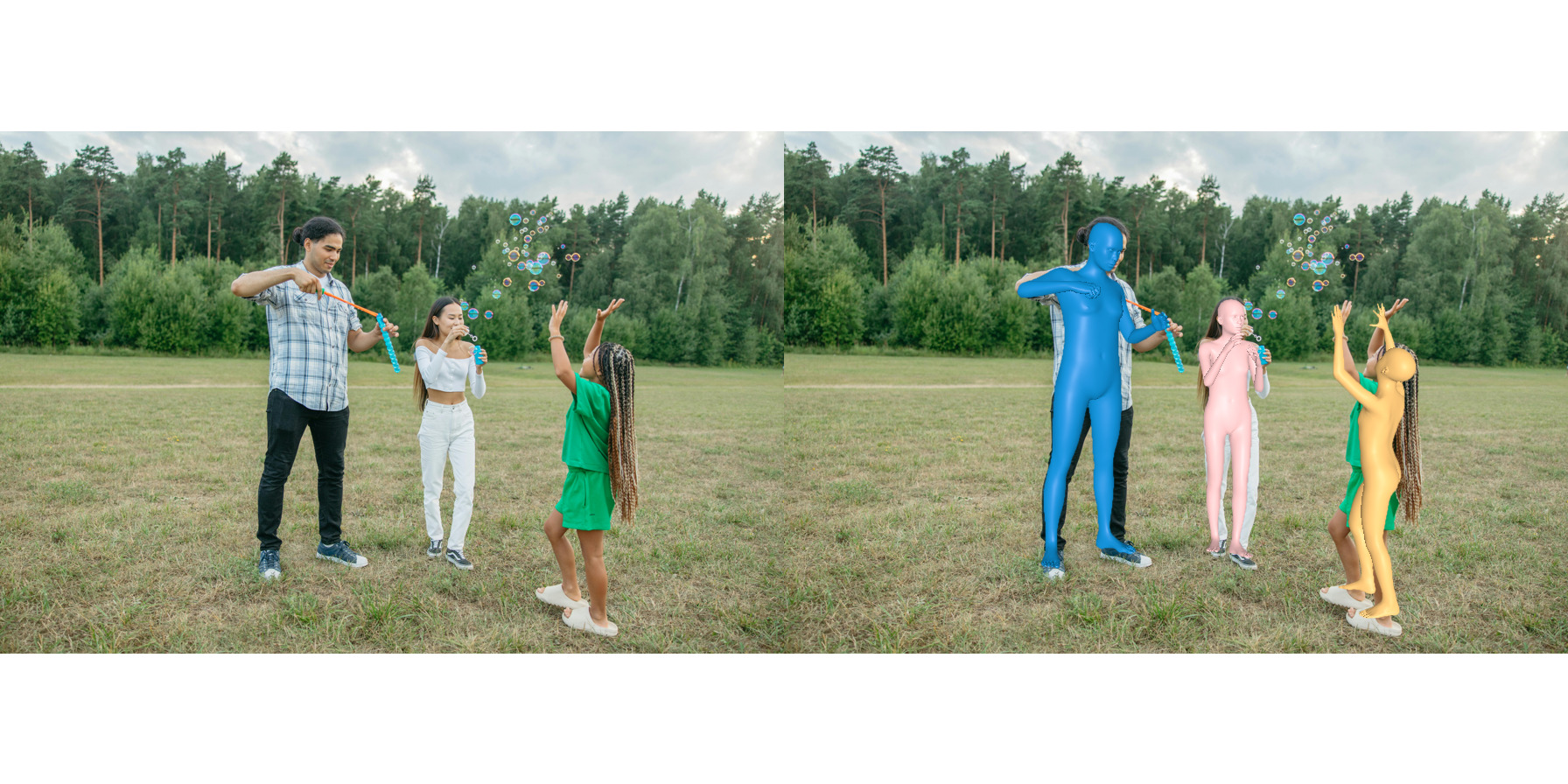}\\
	\end{tabular}
	\vspace{-4mm}
	\caption{\textbf{Qualitative examples} on real-world images sourced from Pexels~\cite{pexels} that contain both adults and children. We observe that our model outputs childlike body shapes for children, in terms of height and build.}
	\vspace{-2mm}
	\label{fig:qualitative}
\end{figure*}

\paragraph{Qualitative results.}
We provide some qualitative examples of results of our approach on real images that contain both adults and children in 
Figure~\ref{fig:qualitative}, showing 
high-quality human mesh recovery results with a single model thanks to \ours's capacity of modeling a wide variety of body shapes.

\section{Conclusion}

We present \ours, a unified, differentiable, and interpretable human body model that does not rely on 3D scan training data, but continuously encodes human shape variations across age, gender, and body type.
Built entirely on open and anthropometric knowledge, \ours bridges the gap between artistic and statistical modeling, offering a transparent alternative to abstract latent spaces.
Despite not being learned from 3D scans, \ours achieves competitive performance when fitting real scan datasets, showing that accurate geometric modeling can be obtained purely from artistic priors and anthropometric calibration.
Using \ours, we generate Anny-One, a large-scale dataset of 780k synthetic images with rich diversity in pose, shape, and scene context.
Experiments show that HMR models trained using \ours achieve 
competitive performance across diverse benchmarks.
By providing a free, interpretable and unified  3D model that covers the full human lifespan, we hope Anny will broaden the ecosystem of human-aware computer vision.

\section{Acknowledgments}

We thank the MakeHuman contributors whose work provided the foundation for Anny, and Timothée Wintz for his help in maintaining the Anny library after this submission.

\bibliographystyle{splncs04}
\bibliography{biblio}

\begin{thebibliography}{10}
\providecommand{\url}[1]{\texttt{#1}}
\providecommand{\urlprefix}{URL }
\providecommand{\doi}[1]{https://doi.org/#1}

\bibitem{sizeusa}
{SizeUSA}. \url{https://www.tc2.com/size-usa.html} (2017)

\bibitem{3dpeople}
{3DPeople}. \url{https://3dpeople.com} ({2020})

\bibitem{humgen3d}
{HumGen3D}. \url{https://www.humgen3d.com/} (2025)

\bibitem{makehuman}
Makehuman. \url{http://www.makehumancommunity.org/} (2025)

\bibitem{mpfb2}
{MPFB2}. \url{https://github.com/makehumancommunity/mpfb2} (2025)

\bibitem{pexels}
Pexels. \url{https://www.pexels.com} (2025)

\bibitem{mixamo}
{Adobe Inc.}: Mixamo. \url{https://www.mixamo.com/} (2025)

\bibitem{andriluka14cvpr}
Andriluka, M., Pishchulin, L., Gehler, P., Schiele, B.: {2D} human pose
  estimation: New benchmark and state of the art analysis. In: CVPR (2014)

\bibitem{armando2024cross}
Armando, M., Galaaoui, S., Baradel, F., Lucas, T., Leroy, V., Br{\'e}gier, R.,
  Weinzaepfel, P., Rogez, G.: Cross-view and cross-pose completion for {3D}
  human understanding. In: CVPR (2024)

\bibitem{multi-hmr2024}
Baradel, F., Armando, M., Galaaoui, S., Br{\'e}gier, R., Weinzaepfel, P.,
  Rogez, G., Lucas, T.: Multi-hmr: Multi-person whole-body human mesh recovery
  in a single shot. In: ECCV (2024)

\bibitem{bedlam}
Black, M.J., Patel, P., Tesch, J., Yang, J.: {BEDLAM}: A synthetic dataset of
  bodies exhibiting detailed lifelike animated motion. In: CVPR (2023)

\bibitem{blender}
{Blender Foundation}: Blender. \url{https://www.blender.org/} (2025)

\bibitem{bogo2016keep}
Bogo, F., Kanazawa, A., Lassner, C., Gehler, P., Romero, J., Black, M.J.: Keep
  it {SMPL}: Automatic estimation of {3D} human pose and shape from a single
  image. In: ECCV (2016)

\bibitem{Bogo:CVPR:2014}
Bogo, F., Romero, J., Loper, M., Black, M.J.: {FAUST}: Dataset and evaluation
  for {3D} mesh registration. In: CVPR (2014)

\bibitem{conman}
Br{\'e}gier, R., Baradel, F., Lucas, T., Galaaoui, S., Armando, M.,
  Weinzaepfel, P., Rogez, G.: Condimen: Conditional multi-person mesh recovery.
  In: CVPR RHOBIN Workshop (2025)

\bibitem{Bulat2017FAW}
Bulat, A., Tzimiropoulos, G.: How far are we from solving the {2D} \& {3D} face
  alignment problem? (and a dataset of 230,000 {3D} facial landmarks). In: ICCV
  (2017)

\bibitem{cai2023smpler}
Cai, Z., Yin, W., Zeng, A., Wei, C., Sun, Q., Yanjun, W., Pang, H.E., Mei, H.,
  Zhang, M., Zhang, L., et~al.: Smpler-x: Scaling up expressive human pose and
  shape estimation. In: NeurIPS (2023)

\bibitem{who2006growth}
{Department of Nutrition for Health and Development}: {WHO} child growth
  standards: length/height-for-age, weight-for-age, weight-for-length,
  weight-for-height and body mass index-for-age: methods and development. Tech.
  rep., World Health Organisation (2006)

\bibitem{feng2021collaborative}
Feng, Y., Choutas, V., Bolkart, T., Tzionas, D., Black, M.J.: Collaborative
  regression of expressive bodies using moderation. In: 3DV (2021)

\bibitem{ferguson2025mhr}
Ferguson, A., Osman, A.A., Bescos, B., Stoll, C., Twigg, C., Lassner, C., Otte,
  D., Vignola, E., Prada, F., Bogo, F., et~al.: Mhr: Momentum human rig. arXiv
  preprint arXiv:2511.15586  (2025)

\bibitem{goel2023humans}
Goel, S., Pavlakos, G., Rajasegaran, J., Kanazawa, A., Malik, J.: Humans in
  4{D}: Reconstructing and tracking humans with transformers. In: ICCV (2023)

\bibitem{guler2018densepose}
G{\"u}ler, R.A., Neverova, N., Kokkinos, I.: Densepose: Dense human pose
  estimation in the wild. In: CVPR (2018)

\bibitem{hesse2018smil}
Hesse, N., Pujades, S., Romero, J., Black, M.J., Bodensteiner, C., Arens, M.,
  Hofmann, U.G., Tacke, U., Hadders-Algra, M., Weinberger, R., Müller-Felber,
  W., Sebastian~Schroeder, A.: Learning an {Infant} {Body} {Model} from
  {RGB}-{D} {Data} for {Accurate} {Full} {Body} {Motion} {Analysis}. In: MICCAI
  (2018)

\bibitem{huang2017towards}
Huang, Y., Bogo, F., Lassner, C., Kanazawa, A., Gehler, P.V., Romero, J.,
  Akhter, I., Black, M.J.: Towards accurate marker-less human shape and pose
  estimation over time. In: 3DV (2017)

\bibitem{ionescu2013human3}
Ionescu, C., Papava, D., Olaru, V., Sminchisescu, C.: {Human3.6M}: Large scale
  datasets and predictive methods for {3D} human sensing in natural
  environments. IEEE Trans. PAMI  (2013)

\bibitem{johnson2010clustered}
Johnson, S., Everingham, M.: Clustered pose and nonlinear appearance models for
  human pose estimation. In: BMVC (2010)

\bibitem{joo2021exemplar}
Joo, H., Neverova, N., Vedaldi, A.: Exemplar fine-tuning for {3D} human model
  fitting towards in-the-wild {3D} human pose estimation. In: 3DV (2021)

\bibitem{cmu_panoptic_studio}
Joo, H., Simon, T., Li, X., Liu, H., Tan, L., Gui, L., Banerjee, S., Godisart,
  T.S., Nabbe, B., Matthews, I., Kanade, T., Nobuhara, S., Sheikh, Y.: Panoptic
  studio: A massively multiview system for social interaction capture. IEEE
  trans. PAMI  (2017)

\bibitem{kanazawa2018end}
Kanazawa, A., Black, M.J., Jacobs, D.W., Malik, J.: End-to-end recovery of
  human shape and pose. In: CVPR (2018)

\bibitem{kaufmann2023emdb}
Kaufmann, M., Song, J., Guo, C., Shen, K., Jiang, T., Tang, C., Z{\'a}rate,
  J.J., Hilliges, O.: {EMDB}: The {E}lectromagnetic {D}atabase of {G}lobal 3{D}
  {H}uman {P}ose and {S}hape in the {W}ild. In: ICCV (2023)

\bibitem{kocabas2020vibe}
Kocabas, M., Athanasiou, N., Black, M.J.: Vibe: Video inference for human body
  pose and shape estimation. In: CVPR (2020)

\bibitem{kolotouros2019learning}
Kolotouros, N., Pavlakos, G., Black, M.J., Daniilidis, K.: Learning to
  reconstruct {3D} human pose and shape via model-fitting in the loop. In: ICCV
  (2019)

\bibitem{lassner2017unite}
Lassner, C., Romero, J., Kiefel, M., Bogo, F., Black, M.J., Gehler, P.V.: Unite
  the people: Closing the loop between {3D} and {2D} human representations. In:
  CVPR (2017)

\bibitem{li2022cliff}
Li, Z., Liu, J., Zhang, Z., Xu, S., Yan, Y.: {CLIFF}: Carrying location
  information in full frames into human pose and shape estimation. In: ECCV
  (2022)

\bibitem{lin2023one}
Lin, J., Zeng, A., Wang, H., Zhang, L., Li, Y.: One-stage {3D} whole-body mesh
  recovery with component aware transformer. In: CVPR (2023)

\bibitem{lin2014microsoft}
Lin, T.Y., Maire, M., Belongie, S., Hays, J., Perona, P., Ramanan, D.,
  Doll{\'a}r, P., Zitnick, C.L.: Microsoft coco: Common objects in context. In:
  ECCV (2014)

\bibitem{loper2015smpl}
Loper, M., Mahmood, N., Romero, J., Pons-Moll, G., Black, M.J.: Smpl: a skinned
  multi-person linear model. In: ACM Trans. Graphics (2015)

\bibitem{warp2022}
Macklin, M.: Warp: A high-performance python framework for gpu simulation and
  graphics  (2022), {NVIDIA GPU Technology Conference (GTC)}

\bibitem{smpl-made-simple}
Mahmood, N., Bolkart, T., Osman, A.A.A., Tesch, J., Tzionas, D., Black, M.J.:
  {SMPL} made {S}imple. CVPR tutorial
  \url{https://smpl-made-simple.is.tue.mpg.de/} (2021)

\bibitem{mahmood2019amass}
Mahmood, N., Ghorbani, N., Troje, N.F., Pons-Moll, G., Black, M.: {AMASS}:
  {Archive} of {Motion} {Capture} {As} {Surface} {Shapes}. In: {ICCV} (2019)

\bibitem{vonMarcard2018}
von Marcard, T., Henschel, R., Black, M., Rosenhahn, B., Pons-Moll, G.:
  Recovering accurate {3D} human pose in the wild using imus and a moving
  camera. In: ECCV (2018)

\bibitem{mono-3dhp2017}
Mehta, D., Rhodin, H., Casas, D., Fua, P., Sotnychenko, O., Xu, W., Theobalt,
  C.: Monocular {3D} human pose estimation in the wild using improved cnn
  supervision. In: 3DV (2017)

\bibitem{moon2022neuralannot}
Moon, G., Choi, H., Lee, K.M.: Neuralannot: Neural annotator for {3D} human
  mesh training sets. In: CVPR (2022)

\bibitem{dinov2}
Oquab, M., Darcet, T., Moutakanni, T., Vo, H., Szafraniec, M., Khalidov, V.,
  Fernandez, P., Haziza, D., Massa, F., El-Nouby, A., et~al.: Dinov2: Learning
  robust visual features without supervision. TMLR  (2024)

\bibitem{osman2022supr}
Osman, A.A.A., Bolkart, T., Tzionas, D., Black, M.J.: {SUPR}: A sparse unified
  part-based human body model. In: ECCV (2022)

\bibitem{osman2020star}
Osman, A.A., Bolkart, T., Black, M.J.: {STAR}: Sparse trained articulated human
  body regressor. In: ECCV (2020)

\bibitem{park2025atlas}
Park, J., Romero, J., Saito, S., Prada, F., Shiratori, T., Xu, Y., Bogo, F.,
  Yu, S.I., Kitani, K., Khirodkar, R.: Atlas: Decoupling skeletal and shape
  parameters for expressive parametric human modeling. In: {ICCV} (2025)

\bibitem{paszke2017automatic}
Paszke, A., Gross, S., Chintala, S., Chanan, G., Yang, E., DeVito, Z., Lin, Z.,
  Desmaison, A., Antiga, L., Lerer, A.: Automatic differentiation in pytorch.
  In: NeurIPS workshop (2017)

\bibitem{patel2025camerahmr}
Patel, P., Black, M.J.: {CameraHMR}: Aligning people with perspective. In: 3DV
  (2025)

\bibitem{agora}
Patel, P., Huang, C.H.P., Tesch, J., Hoffmann, D.T., Tripathi, S., Black, M.J.:
  {AGORA}: Avatars in geography optimized for regression analysis. In: CVPR
  (2021)

\bibitem{pavlakos2019smplx}
Pavlakos, G., Choutas, V., Ghorbani, N., Bolkart, T., Osman, A.A., Tzionas, D.,
  Black, M.J.: Expressive body capture: {3D} hands, face, and body from a
  single image. In: CVPR (2019)

\bibitem{infinigen2024indoors}
Raistrick, A., Mei, L., Kayan, K., Yan, D., Zuo, Y., Han, B., Wen, H., Parakh,
  M., Alexandropoulos, S., Lipson, L., Ma, Z., Deng, J.: Infinigen indoors:
  Photorealistic indoor scenes using procedural generation. In: CVPR (2024)

\bibitem{robinette2002caesar}
Robinette, K.M., Blackwell, S., Daanen, H., Boehmer, M., Fleming, S.: Civilian
  {American} and {European} {Surface} {Anthropometry} {Resource} ({CAESAR}),
  {Final} {Report}. {Volume} 1. {Summary}:. Tech. rep., Defense Technical
  Information Center (2002)

\bibitem{rodriguez2020height}
Rodriguez-Martinez, A., Zhou, B., Sophiea, M.K., Bentham, J., Paciorek, C.J.,
  Iurilli, M.L., Carrillo-Larco, R.M., Bennett, J.E., Di~Cesare, M., Taddei,
  C., et~al.: Height and body-mass index trajectories of school-aged children
  and adolescents from 1985 to 2019 in 200 countries and territories: a pooled
  analysis of 2181 population-based studies with 65 million participants. The
  Lancet  (2020)

\bibitem{saint20183dbodytex}
Saint, A., Ahmed, E., Shabayek, A.E.R., Cherenkova, K., Gusev, G., Aouada, D.,
  Ottersten, B.: {3DBodyTex}: {Textured} {3D} {Body} {Dataset}. In: {3DV}
  (2018)

\bibitem{shin2024wham}
Shin, S., Kim, J., Halilaj, E., Black, M.J.: Wham: Reconstructing
  world-grounded humans with accurate {3D} motion. In: CVPR (2024)

\bibitem{su2025sat}
Su, C., Ma, X., Su, J., Wang, Y.: Sat-hmr: Real-time multi-person {3D} mesh
  estimation via scale-adaptive tokens. In: CVPR (2025)

\bibitem{sun2024aios}
Sun, Q., Wang, Y., Zeng, A., Yin, W., Wei, C., Wang, W., Mei, H., Leung, C.S.,
  Liu, Z., Yang, L., et~al.: Aios: All-in-one-stage expressive human pose and
  shape estimation. In: CVPR (2024)

\bibitem{sun2021monocular}
Sun, Y., Bao, Q., Liu, W., Fu, Y., Black, M.J., Mei, T.: Monocular, one-stage,
  regression of multiple {3D} people. In: ICCV (2021)

\bibitem{sun2022putting}
Sun, Y., Liu, W., Bao, Q., Fu, Y., Mei, T., Black, M.J.: Putting people in
  their place: Monocular regression of {3D} people in depth. In: CVPR (2022)

\bibitem{taheri2020grab}
Taheri, O., Ghorbani, N., Black, M.J., Tzionas, D.: {GRAB}: {A} {Dataset} of
  {Whole}-{Body} {Human} {Grasping} of {Objects}. In: {ECCV} (2020)

\bibitem{tesch2025bedlam2}
Tesch, J., Becherini, G., Achar, P., Yiannakidis, A., Kocabas, M., Patel, P.,
  Black, M.J.: {BEDLAM}2.0: Synthetic humans and cameras in motion. In: NeurIPS
  (2025)

\bibitem{metahuman}
{Unreal Engine}: {MetaHuman}. \url{https://www.unrealengine.com/fr/metahuman}
  (2025)

\bibitem{varol2017learning}
Varol, G., Romero, J., Martin, X., Mahmood, N., Black, M.J., Laptev, I.,
  Schmid, C.: Learning from synthetic humans. In: CVPR (2017)

\bibitem{wang2024blade}
Wang, S., Li, J., Li, T., Yuan, Y., Fuchs, H., De~Mello, S., Nagano, K.,
  Stengel, M.: {BLADE}: {S}ingle-view {B}ody {M}esh {L}earning through
  {A}ccurate {D}epth {E}stimation. arXiv preprint arXiv:2412.08640  (2024)

\bibitem{wang2023zolly}
Wang, W., Ge, Y., Mei, H., Cai, Z., Sun, Q., Wang, Y., Shen, C., Yang, L.,
  Komura, T.: Zolly: Zoom focal length correctly for perspective-distorted
  human mesh reconstruction. In: ICCV (2023)

\bibitem{xu2020ghum}
Xu, H., Bazavan, E.G., Zanfir, A., Freeman, W.T., Sukthankar, R., Sminchisescu,
  C.: {GHUM} \& {GHUML}: Generative {3D} human shape and articulated pose
  models. In: CVPR (2020)

\bibitem{yin2024whac}
Yin, W., Cai, Z., Wang, R., Wang, F., Wei, C., Mei, H., Xiao, W., Yang, Z.,
  Sun, Q., Yamashita, A., et~al.: {WHAC}: World-grounded humans and cameras.
  In: ECCV (2024)

\bibitem{yin2025smplest}
Yin, W., Cai, Z., Wang, R., Zeng, A., Wei, C., Sun, Q., Mei, H., Wang, Y.,
  Pang, H.E., Zhang, M., Zhang, L., Loy, C.C., Yamashita, A., Yang, L., Liu,
  Z.: {SMPLest-X}: Ultimate scaling for expressive human pose and shape
  estimation. IEEE Trans. PAMI  (2025)

\bibitem{yin2023hi4d}
Yin, Y., Guo, C., Kaufmann, M., Zarate, J., Song, J., Hilliges, O.: Hi4d: 4d
  instance segmentation of close human interaction. In: CVPR (2023)

\bibitem{zhang2021pymaf}
Zhang, H., Tian, Y., Zhou, X., Ouyang, W., Liu, Y., Wang, L., Sun, Z.: {PyMAF}:
  {3D} human pose and shape regression with pyramidal mesh alignment feedback
  loop. In: ICCV (2021)

\end{thebibliography}
\end{document}